\documentclass[10.5pt]{article}
\usepackage{graphicx} 
\usepackage[margin=1in]{geometry}
\usepackage{amsmath}
\usepackage{amssymb}
\usepackage[T1]{fontenc}
\usepackage{lmodern}

\usepackage[backend=biber, style=numeric, sorting=none]{biblatex}
\renewbibmacro*{finentry}{%
  \finentry%
  \par\smallskip
  \begingroup\small
  \endgroup\par}
\renewbibmacro*{in:}{}

\newcommand {\bx}{\mathbf{x}}

\newcommand {\bC}{\mathbf{C}}
\newcommand {\bm}{\mathbf{m}}
\newcommand {\bH}{\mathbf{H}}
\newcommand {\bth}{\boldsymbol{\theta}}
\newcommand {\bSigma}{\boldsymbol{\Sigma}}
\newcommand {\bv}{\mathbf{v}}
\newcommand {\bF}{\mathbf{F}}

\title{Stochastic learning flow: moment equations for stochastic gradient descent at finite step size}
\author{
  Christian Schmid\textsuperscript{1,2}, 
  James M Murray\textsuperscript{1}\footnote{Contact: \texttt{jmurray9@uoregon.edu}}\\[0.5ex]
  \textsuperscript{1}Institute of Neuroscience, University of Oregon\\
  \textsuperscript{2}Center for Neural Science, New York University%
}

\begin{document}

\maketitle

\begin{abstract}
Analytical theories of learning in deep networks often assume gradient flow, discarding the finite step size and batch noise of stochastic gradient descent (SGD). Starting from the discrete update, we derive continuous-time equations for the mean and covariance of the parameter distribution under a Gaussian closure for a general stochastic learning, to first order in the learning rate $\eta$. Applying these stochastic learning flow (SLF) equations to SGD, we show that the mean descends an effective loss regularized by the covariance and loss curvature, while the covariance, unlike in stochastic differential equation approximations, becomes unstable at curvature $2/\eta$. In SGD ensembles of linear and nonlinear networks, SLF predicts learning curves and mean trajectories more accurately than gradient flow and captures the covariance dynamics, failing where the parameter distribution becomes strongly non-Gaussian. In the full-batch limit, the edge of stability corresponds to a stationary point of the coupled sharpness-covariance dynamics, recovering self-stabilization and central flow, and batch-to-batch curvature fluctuations lower this threshold under SGD.
\end{abstract}

\section{Introduction}
Many of the most successful analytical theories of learning in deep networks, including exact solutions for deep linear networks~\cite{saxe2013exact,saxe2019mathematical} and the neural tangent kernel~\cite{jacot2018neural}, are formulated in the continuous-time limit of gradient flow. This limit makes the tools of dynamical systems theory available, but at the cost of discarding the finite step size and the sampling noise of stochastic, iterative algorithms such as stochastic gradient descent (SGD). These are not small corrections in practice. The learning rate and batch size change which solutions training finds and how well they generalize~\cite{keskar2017large,jastrzkebski2017three,smith2018bayesian,li2019towards}, large learning rates produce qualitatively new dynamics~\cite{lewkowycz2020large,even2023s}, and networks trained with gradient descent typically operate at the edge of stability, where the product of the learning rate and the largest Hessian eigenvalue is close to $2$, far from the regime in which gradient flow applies~\cite{cohen2021gradient}.

A widely used approach approximates SGD by a stochastic differential equation (SDE), in which gradient flow is augmented by a diffusion term matched to the covariance of the minibatch gradient noise~\cite{mandt2016variational,jastrzkebski2017three,chaudhari2018stochastic,li2017stochastic}. Higher-order versions add deterministic finite-step corrections to the drift~\cite{li2017stochastic,barrett2020implicit,smith2021origin,kunin2023limiting}, but as expansions in the product of the learning rate and the curvature, which break down where that product is of order one. In particular, these descriptions contain no instability at curvature $2/\eta$, and SDE surrogates of SGD do not stabilize where SGD does~\cite{andreyev2024edge}. 

Conversely, the edge of stability has been analyzed with largely deterministic theories~\cite{damian2023self,cohen2025understanding}, which recent work extends to SGD through local stability analyses \cite{wu2018sgd,mulayoff2024exact} or additive noise~\cite{liao2026sgd}. 
In high-dimensional linear \cite{paquette2021sgd,paquette2025homogenization,bordelon2022learning} and generalized-linear \cite{collins2024hitting} models, exact deterministic equations for the risk under SGD at finite step size and batch size are available , but they rely on the quadratic structure of the loss.
Ensembles of stochastic learners have also been described directly at the level of the parameter distribution. 
Kramers–Moyal expansions of the master equation for on-line learning at constant learning rate have long been used to describe the parameter distribution of stochastic learners, focusing on Fokker–Planck dynamics and their equilibria \cite{leen1992weight} or on moment equations with random update times \cite{heskes1991learning}.
Working instead from the discrete update, Yaida \cite{yaida2019fluctuation} derived exact fluctuation–dissipation relations at stationarity, and Liu et al. \cite{liu2021noise} obtained the exact stationary covariance of SGD near a quadratic minimum.
What is missing is a continuous-time description of the parameter distribution that retains the deterministic effects of finite step size and the stochastic effects of minibatch sampling together, without expanding in the product of learning rate and curvature.

To derive such a description, we start from the discrete update rule and obtain, for general architectures and update rules, evolution equations for the mean and covariance of the parameter distribution $p(\bth,t)$ that retain all terms at first order in the learning rate. In these \textit{stochastic learning flow} (SLF) equations, the parameter covariance is a dynamical variable on the same footing as the mean. Its noise-driven part coincides with the diffusion of SDE models, but additional finite-step terms qualitatively change its dynamics. 
For SGD, the mean descends an effective loss that includes an implicit regularization term set by the covariance and the loss curvature, which provides a mechanism for escape from saddle points and sharp minima.

We test the theory against ensembles of networks trained with SGD. In a deep linear network, SLF predicts the learning curves and parameter updates far more accurately than gradient flow at large learning rates and small batch sizes. In a nonlinear network trained on binary MNIST, the covariance correction improves the prediction of the mean by an amount that grows with the batch noise, and the integrated covariance matches the measured one. In both cases the description fails where the parameter distribution becomes strongly non-Gaussian. Finally, the SLF equations contain the edge-of-stability threshold and central flow as limiting cases, and show how fluctuations of the curvature across batches lower the threshold under SGD.

\section{Results}
Consider an artificial neural network with parameters $\bth$ and update equation
\begin{equation}
    \label{eq:20260420a}
    \theta_k(t + \delta t) = \theta_k(t) + \eta f_k(\bth(t)),
\end{equation}
where $\eta$ is a learning rate, and $f_k$ is stochastic. For example, in stochastic gradient descent, we have $\mathbf{f} = -\nabla L$, where $L$ is the loss function computed over a batch of data, so that the stochasticity arises from the random draws of data in each batch. In the case of reinforcement learning, $\mathbf{f}$ may depend on exploratory noise as well.

In the sections that follow, we will first derive continuous-time moment equations describing the noisy evolution of the parameters during learning for general update equations \eqref{eq:20260420a}, then specialize and interpret the equations in the particular case of SGD.
To validate the theory and establish its limitations, we will test these equations against empirical ensembles of SGD learners in linear and nonlinear networks.
Finally, we will show how these generalized equations account for EoS dynamics.

\subsection{General dynamical equations for noisy parameter evolution}
\label{sec:general_derivation}
Beginning from \eqref{eq:20260420a}, our goal will be to derive an equation characterizing the time evolution of the distribution $p(\theta, t)$ of network weights during training. 
Our approach is based on the Kramers–Moyal expansion of the master equation for the discrete update \cite{risken1989fokker,kramers1940brownian,moyal1949stochastic}. It follows earlier work that described ensembles of on-line learning processes at constant learning rate with master and Fokker–Planck equations \cite{heskes1991learning,leen1992weight},
and that we previously applied at leading order to learning in single-layer networks \cite{murray2020remembrance,schmid2024dynamics}. Related moment relations, obtained from the stationarity of the discrete-time update, yield fluctuation–dissipation relations for SGD \cite{yaida2019fluctuation}.
To do this, we begin with the following ``master equation", which describes the evolution of this distribution over a single learning step:
\begin{equation}
    \label{eq:20260420b}
    p(\bth, t + \delta t) = \int d \delta\bth ~ 
    p(\bth, t + \delta t | \bth - \delta\bth, t)
    p(\bth - \delta\bth, t).
\end{equation}
Assuming the step size to be small, the left-hand side of this equation can be expanded as
\begin{equation}
    \label{eq:20260420c}
    p(\bth, t + \delta t) = 
    p(\bth, t)
    + \delta t \partial_t p(\bth, t)
    + \frac12 \delta t^2 \partial_t^2 p(\bth, t)
    + \ldots,
\end{equation}
and the integrand from the right-hand side as
\begin{equation}
\begin{split}
    \label{eq:20260420d}
    p(\bth + \delta\bth - \delta\bth, t + \delta t | \bth - \delta\bth, t)
    p(\bth - \delta\bth, t)
    &= p(\bth + \delta\bth, t + \delta t | \bth, t) p(\bth, t) \\
    \quad & - \sum_i \delta\theta_i \frac{\partial}{\partial \theta_i}
    [p(\bth + \delta\bth, t + \delta t | \bth, t) p(\bth, t)] \\
    \quad & + \frac12 \sum_{ij} \delta\theta_i \delta\theta_j
    \frac{\partial^2}{\partial \theta_i \partial \theta_j}
    [p(\bth + \delta\bth, t + \delta t | \bth, t) p(\bth, t)]
    + \ldots
\end{split}
\end{equation}
Setting the two sides of the equation equal to one another then gives
\begin{equation}
\label{eq:20260420e}
    \left( \delta t \partial_t + \frac12 \delta t^2 \partial_t^2 \right)
    p(\bth, t)
    = - \sum_i \frac{\partial}{\partial \theta_i}
    [\alpha_i(\bth) p(\bth, t)]
    + \frac12 \sum_{ij} 
    \frac{\partial^2}{\partial \theta_i \partial \theta_j}
    [\beta_{ij} (\bth) p(\bth, t)]
    + O(\delta t^3, \eta^3),
\end{equation}
where
\begin{equation}
\label{eq:20260420f}
    \alpha_i(\bth) = \int d\delta\bth ~ \delta \theta_i
    p(\bth + \delta\bth, t + \delta t | \bth, t)
\end{equation}
and
\begin{equation}
\label{eq:20260420g}
    \beta_{ij}(\bth) = \int d\delta\bth ~ \delta \theta_i \delta \theta_j
    p(\bth + \delta\bth, t + \delta t | \bth, t).
\end{equation}
Using \eqref{eq:20260420a}, these last two equations are
\begin{align}
\label{eq:20260420h}
    \alpha_i(\bth) &= \eta \langle f_i (\bth) \rangle, \\
    \beta_{ij} (\bth) &= \eta^2 \langle f_i (\bth) f_j (\bth) \rangle,
\end{align}
where $\langle \cdot \rangle$ denotes an average over the stochasticity of a single update at fixed $\bth$ (for SGD, over the draw of a batch). 
(Note that, in the case of reinforcement learning, this would additionally include averaging over exploratory noise in the network.)
Below we write $\langle \cdot \rangle_{\bth}$ for an average over $p(\bth, t)$, and $\mathrm{Cov}(\cdot,\cdot)$ and $\mathrm{Var}(\cdot)$ for covariances and variances with respect to $\langle \cdot \rangle$; 
for example, $\mathrm{Cov}(f_k,f_l)=\langle f_k f_l\rangle-\langle f_k\rangle\langle f_l\rangle$. For finite ensembles of trained networks, $\mathbb{E}[\bth]$ and $\mathrm{Cov}(\bth)$ denote the sample mean and covariance over replicas.

Thus, dropping the higher-order terms, \eqref{eq:20260420e} gives the following Fokker-Planck-like description of the dynamics of the parameter vector:
\begin{equation}
\label{eq:20260420i}
    \left( \delta t \partial_t + \frac12 \delta t^2 \partial_t^2 \right)
    p(\bth, t)
    = - \eta \sum_i \frac{\partial}{\partial \theta_i}
    [\langle f_i(\bth) \rangle p(\bth, t)]
    + \frac{\eta^2}{2} \sum_{ij} 
    \frac{\partial^2}{\partial \theta_i \partial \theta_j}
    [\langle f_i(\bth) f_j(\bth) \rangle p(\bth, t)].
\end{equation}
However, note that we have kept the $O(\delta t^2)$ term in the expansion, which is nonstandard in the ordinary Fokker-Planck derivation but, as we will see below, is crucial for accounting for the effects induced by finite step size.

Related to this, we can see that the two terms on the right-hand side of \eqref{eq:20260420i} scale with different powers of the small parameter $\eta$. 
This arises from the fact that both the deterministic and stochastic parts of the update equation scale $\sim\eta$ in \eqref{eq:20260420a}, which is in contrast to classical theories such as Brownian motion, where the two terms are taken to scale as $\sim \delta t$ and $\sim \sqrt{\delta t}$, respectively, leading to an equation in which the drift and diffusion terms are both of the same order in $\delta t$.
Hence, given the update equation \eqref{eq:20260420a} that defines iterative learning, the next-to-leading order expansion in \eqref{eq:20260420i} is the simplest consistent evolution equation for $p(\bth, t)$ that accounts for noise-related effects arising from input (or output) noise and finite step size.

Rather than attempting to characterize the evolution of the full distribution $p(\bth, t)$, we can gain more insight by analyzing the time evolution of the first two moments. To do this, we multiply both sides of \eqref{eq:20260420i} by powers of $\bth$ and integrate. For the first moment, this gives
\begin{equation}
\begin{split}
    \label{eq:20260420j}
    \left( \delta t \partial_t + \frac12 \delta t^2 \partial_t^2 \right)
    \int d\bth ~\theta_k p(\bth, t)
    &= - \eta \sum_i \int d\bth 
    ~\theta_k \frac{\partial}{\partial \theta_i}
    [\langle f_i(\bth) \rangle p(\bth, t)] \\
    & \quad + \frac{\eta^2}{2} \sum_{ij} \int d\bth ~ \theta_k
    \frac{\partial^2}{\partial \theta_i \partial \theta_j}
    [\langle f_i(\bth) f_j(\bth) \rangle p(\bth, t)].
\end{split}
\end{equation}
Integrating this by parts then gives
\begin{equation}
    \label{eq:20260420k}
    \left( \delta t \partial_t + \frac12 \delta t^2 \partial_t^2 \right)
    \langle \theta_k \rangle_{\bth} 
    = \eta \langle\langle f_k(\bth) \rangle\rangle_{\bth}.
\end{equation}
Similarly, we can multiply by $\theta_k \theta_l$ and integrate by parts to obtain
\begin{equation}
    \label{eq:20260420l}
    \left( \delta t \partial_t + \frac12 \delta t^2 \partial_t^2 \right)
    \langle \theta_k \theta_l \rangle_{\bth} 
    = \eta \langle \theta_k \langle f_l(\bth) \rangle\rangle_{\bth} 
    + \eta \langle \theta_l \langle f_k(\bth) \rangle\rangle_{\bth} 
    + \eta^2 \langle\langle f_k(\bth) f_l(\bth) \rangle\rangle_{\bth}.
\end{equation}
We will comment on the limitations associated with this choice in discussing our empirical results below.

As is typically done in continuous-time formulations of gradient descent, we can now let $\delta t = \eta$. Doing this and defining 
\begin{equation}
\label{eq:20260420m}
D_t = \partial_t + \frac{\eta}{2} \partial_t^2,
\end{equation}
Equations \eqref{eq:20260420k} and \eqref{eq:20260420l} become
\begin{equation}
    \label{eq:20260420n}
    D_t \langle \theta_k \rangle_{\bth} 
    = \langle\langle f_k(\bth) \rangle\rangle_{\bth}
\end{equation}
and
\begin{equation}
    \label{eq:20260420o}
    D_t \langle \theta_k \theta_l \rangle_{\bth} 
    = \langle \theta_k \langle f_l(\bth) \rangle\rangle_{\bth} 
    + \langle \theta_l \langle f_k(\bth) \rangle\rangle_{\bth} 
    + \eta \langle\langle f_k(\bth) f_l(\bth) \rangle\rangle_{\bth}.
\end{equation}
While, by setting $\delta t = \eta$ in our equations, we do assume that $\eta$ is small, we do not take the limit $\eta \to 0$ as in the commonly studied case of gradient flow. Rather, we assume only that $\eta \ll 1$, such that terms $\sim O(\eta)$ in the above equations are interesting, while terms $\sim O(\eta^2)$ can be neglected. At this point we can see the justification for keeping the $O(\delta t^2)$ term on the left-hand side of \eqref{eq:20260420i}: given the definition of the update equation, this scheme keeps terms up to the same order on both sides of our dynamics equations.

Combining \eqref{eq:20260420n} and \eqref{eq:20260420o}, we can obtain an equation for the parameter covariance, which gives the following closed pair of dynamical equations for the first two moments:
\begin{equation}
\begin{split}
\label{eq:20260420p}
    D_t \langle \theta_k \rangle_{\bth} 
    &= \langle\langle f_k(\bth) \rangle\rangle_{\bth}, \\
    D_t (\langle \theta_k \theta_l \rangle_{\bth}
    - \langle \theta_k \rangle_{\bth}
    \langle \theta_l \rangle_{\bth})
    &= \langle (\theta_k - \langle \theta_k \rangle_{\bth})
    \langle f_l(\bth) \rangle \rangle_{\bth}
    + \langle (\theta_l - \langle \theta_l \rangle_{\bth})
    \langle f_k(\bth) \rangle \rangle_{\bth} \\
    &\quad + \eta \left[ 
    \langle \langle f_k(\bth) f_l(\bth) \rangle \rangle_{\bth}
    - \langle \langle f_k(\bth) \rangle \rangle_{\bth}
    \langle \langle f_l(\bth) \rangle \rangle_{\bth}
    \right] .
\end{split}
\end{equation}
Next, we can expand in the fluctuations of the parameters about their mean values:
\begin{equation}
\begin{split}
\label{eq:20260420q}
    f_k(\bth) &= f_k(\langle \bth \rangle_{\bth}
    + \bth - \langle \bth \rangle_{\bth}) \\
    &= f_k(\langle \bth \rangle_{\bth}) 
    + \sum_i (\theta_i - \langle \theta_i \rangle_{\bth})
    \partial_i f_k(\langle \bth \rangle_{\bth})
    + \frac12 \sum_{ij} (\theta_i - \langle \theta_i \rangle_{\bth})
    (\theta_j - \langle \theta_j \rangle_{\bth})
    \partial_i \partial_j f_k(\langle \bth \rangle_{\bth})
    + O([\bth - \langle \bth \rangle_{\bth}]^3),
\end{split}
\end{equation}
where we have defined $\partial_i = \tfrac{\partial}{\partial\langle\bth\rangle_{\bth}}$. 
Truncating at second order in $\delta\bth$ amounts to a Gaussian closure, whereby third and higher cumulants of $p(\bth, t)$ are neglected, as are terms of $O(\bC^2)$. This requires fluctuations to be small on the scale over which $\mathbf{f}$ varies. Sections \ref{sec:DLN} and \ref{sec:nonlinear} examine where this fails.

If we also define $f_k = f_k(\langle\bth\rangle_{\bth})$, $m_k = \langle\bth_k\rangle_{\bth}$, and $C_{kl} = \langle\bth_k\bth_l\rangle_{\bth} - \langle\bth_k\rangle_{\bth} \langle\bth_l\rangle_{\bth}$, the evolution equations \eqref{eq:20260420p} become
\begin{equation}
\label{eq:20260420r}
    D_t m_k = \langle f_k \rangle
    + \frac{1}{2}\sum_{ij}C_{ij}\,\partial_i\partial_j\langle f_k \rangle
\end{equation}
and
\begin{equation}
\begin{split}
\label{eq:20260420s}
    D_t C_{kl} 
    &= \sum_i C_{ki} \partial_i \langle f_l \rangle
    + \sum_i C_{li} \partial_i \langle f_k \rangle 
    + \eta \bigg\{
    \langle f_k f_l \rangle - \langle f_k \rangle \langle f_l \rangle \\
    & + \frac{1}{2} \sum_{ij} C_{ij} 
    \bigg[
    \langle f_k \partial_i \partial_j f_l \rangle 
    - \langle f_k \rangle \partial_i \partial_j \langle f_l \rangle
    + \langle f_l \partial_i \partial_j f_k \rangle
    - \langle f_l \rangle \partial_i \partial_j \langle f_k \rangle
    + 2 \langle (\partial_i f_k) (\partial_j f_l) \rangle
    \bigg]
    \bigg\}.
\end{split}
\end{equation}
These two equations describe the coupled flow of the parameter mean and covariance for a general stochastic learning rule.

At leading order, near a fixed point and with Poisson-distributed update times, \eqref{eq:20260420r}-\eqref{eq:20260420s} reduce to the moment equations of Ref.~\cite{heskes1991learning}. 
Two features distinguish the results here from that earlier work, however. 
First, the $O(\eta)$ terms in \eqref{eq:20260420s}, including $\eta\langle(\partial_i f_k)C_{ij}(\partial_j f_l)\rangle$, are retained. 
Second, rather than being random, the update times are fixed, which enters through $D_t$: its $\partial_t^2$ term converts $\eta\langle\langle f_k f_l\rangle\rangle_\theta$ into the covariance $\eta[\langle\langle f_k f_l\rangle\rangle_\theta - \langle\langle f_k\rangle\rangle_\theta\langle\langle f_l\rangle\rangle_\theta]$ in \eqref{eq:20260420p}. 
With Poisson timing, by contrast, even full-batch gradient descent would acquire covariance from the randomness of the update times.

The general results \eqref{eq:20260420r} and \eqref{eq:20260420s} enable us to analyze the dynamics of learning for arbitrary architectures (without assuming infinite-width networks) and learning rules, where the noise arises both from finite step size and from noise in the update rule itself due to sampling noise and/or exploratory noise. 
In the language of physics, \eqref{eq:20260420r} and \eqref{eq:20260420s} describe the motion of very light particles subject to large viscous drag forces as well as externally applied forces (the right-hand sides of the two equations). 
The first effect of $\eta > 0$ is to endow these particles with a small mass, $\tfrac{\eta}{2}$. 
As we will analyze in detail in the following section, the other, more consequential effect is to subject the particles to additional forces that can qualitatively change the ways in which their trajectories evolve.

\subsection{Evolution equations for stochastic gradient descent}
\label{sec:SGD}
In stochastic gradient descent (SGD), the update rule is given by $f_k = - \partial_k L$, where $L$ is the loss function defined from a batch of data (and hence is a stochastic variable). If we additionally define $H_{ij} = \partial_i \partial_j L$ as the Hessian of the loss function, the dynamical equations \eqref{eq:20260420r} and \eqref{eq:20260420s} become
\begin{equation}
\label{eq:20260420t}
    D_t m_k = \langle f_k \rangle 
    - \frac12 \sum_{ij} C_{ij} \partial_k \langle H_{ij} \rangle
\end{equation}
and
\begin{equation}
\begin{split}
\label{eq:20260420u}
    D_t C_{kl} &= - \sum_i (C_{ki} \langle H_{il} \rangle
    + \langle H_{ki} \rangle C_{il})
    + \eta \sum_{ij} C_{ij}
    \langle H_{ik} H_{jl} \rangle\\
    & \quad + \eta \bigg\{
    \mathrm{Cov}(f_k, f_l)
    - \frac12 \sum_{ij} C_{ij} \left[
    \mathrm{Cov}(f_k, \partial_l H_{ij})
    + \mathrm{Cov}(f_l, \partial_k H_{ij})
    \right] \bigg\}.
\end{split}
\end{equation}
Equations \eqref{eq:20260420t} and \eqref{eq:20260420u} show that nontrivial parameter dynamics during learning arise not only from the gradient of the loss function, as captured by the gradient-flow equation $\partial_t m_k = \langle f_k \rangle$, but also through its local curvature, which is encoded in the Hessian $\bH$.
To understand the dynamics of $\bC$, we can separately consider the effects of each of the terms on the right hand side of \eqref{eq:20260420u}. 

Because $\bC$ is a dynamical variable independent of $\bm$, $\sum_{ij} C_{ij} \partial_k \langle H_{ij} \rangle = \partial_k \mathrm{Tr}(C \langle H \rangle)$, with the gradient taken at fixed $C$, so that \eqref{eq:20260420t} and \eqref{eq:20260420u} can be written more compactly as
\begin{equation}
\label{eq:20260501a}
D_t \bm = \langle \mathbf{f} \rangle
- \frac12 \nabla \mathrm{Tr} 
(\bC \langle \bH \rangle)
\end{equation}
and
\begin{equation}
\begin{split}
\label{eq:20260501b}
D_t \bC &= - (\bC \langle \bH \rangle
+ \langle \bH \rangle \bC)
+ \eta \langle \bH \bC \bH \rangle \\
& \quad + \eta \left\{
\mathrm{Cov}(\mathbf{f})
- \frac12 \left[ 
\mathrm{Cov}(\mathbf{f}, 
\nabla \mathrm{Tr}(\bC \bH))
+ \mathrm{Cov}(\nabla \mathrm{Tr}
(\bC \bH), \mathbf{f})
\right] \right\}
\end{split}
\end{equation}
Equation \eqref{eq:20260501a} shows that the parameters evolve under SGD via gradient descent on an effective loss function:
\begin{equation}
\label{eq:20260501c}
L^\mathrm{eff} (\bm) = \langle L (\bm) \rangle
+ \frac12 \mathrm{Tr}[\bC(t) \langle \bH(\bm) \rangle].
\end{equation}
The \textit{dynamic} regularization term in this loss function modifies gradient descent by pushing the solution away from regions of large positive curvature and toward flat valleys in the loss landscape.
The details of its effects, however, depend crucially on the dynamics of $\bC(t)$ arising from \eqref{eq:20260501b}, where these dynamics are themselves driven by the curvature of the loss landscape.

The first line of \eqref{eq:20260501b} is driven by the geometry of the loss landscape.
To see this clearly, consider the full-batch limit, in which $\langle \bH \rangle = \bH$, and the second line of \eqref{eq:20260501b} vanishes.
Written in the basis in which $\bH$ is diagonal, we have $\bH = \mathbf{S} \boldsymbol{\Lambda} \mathbf{S}^\top$, where $\boldsymbol{\Lambda} = \mathrm{diag}(\lambda_1, \lambda_2, \ldots)$, and $\tilde{\bC} = \mathbf{S}^\top \bC \mathbf{S}$.
Equation \eqref{eq:20260501b} then becomes
\begin{equation}
\label{eq:20260426a}
    D_t \tilde{C}_{kl} 
    = (\eta \lambda_k \lambda_l - \lambda_k - \lambda_l)
    \tilde{C}_{kl}.
\end{equation}
For a quadratic loss, \eqref{eq:20260426a} is exact rather than truncated: its right-hand side equals $[(1-\eta\lambda_k)(1-\eta\lambda_l)-1]\tilde{C}_{kl}/\eta$, the one-step change of $\tilde{C}_{kl}$ under gradient descent divided by the step size. The only approximation is replacing this finite difference with $D_t$. The instability at $\eta\lambda=2$ is therefore located exactly, while growth and decay rates near it are approximate when $\eta\lambda$ is not small.
In the limit $\eta \to 0$, \eqref{eq:20260426a} shows that modes of $\bC$ aligned with directions of positive loss curvature are pushed to exponentially shrink, while those aligned with directions of negative loss curvature are pushed to exponentially grow, with the eigenvalues of $\bH$ controlling the rate of convergence or divergence.
In the case where $\eta \to 0$ and the loss is convex, we can see that $\bC \to 0$, and our equations reduce to gradient flow, with $\partial_t m_k = \langle f_k \rangle$. 
However, if the loss is non-convex, we see that $\bC$ is amplified in regions of negative curvature even in the limit $\eta \to 0$, which, as noted in previous work \cite{dauphin2014identifying,saxe2013exact}, means that nearby trajectories diverge transiently along negative-curvature directions encountered during training.

While the $O(\eta^0)$ term pulls the parameters down into valleys, \eqref{eq:20260426a} shows that introducing finite step size makes these valleys unstable once they become too narrow, providing a destabilizing mechanism of escape from saddle points and local minima.
In particular, \eqref{eq:20260426a} shows that the dynamics near a local minimum become unstable if $\eta > \tfrac{2}{\lambda_\mathrm{max}}$, providing a mechanism for the network to avoid converging in sharp local minima.
This is the \textit{edge of stability} criterion \cite{lecun1991eigenvalues,cohen2021gradient}: the parameter covariance along any particular direction is stable if and only if the corresponding Hessian eigenvalue satisfies $\lambda < \tfrac{2}{\eta}$.
We analyze this more fully in Section \ref{sec:EoS} below.
Together, the terms in the first line of \eqref{eq:20260501b} describe a competition between terms that push and pull the parameters toward and away from loss minima.

\begin{figure}
    \centering
    \includegraphics[width=0.8\linewidth]{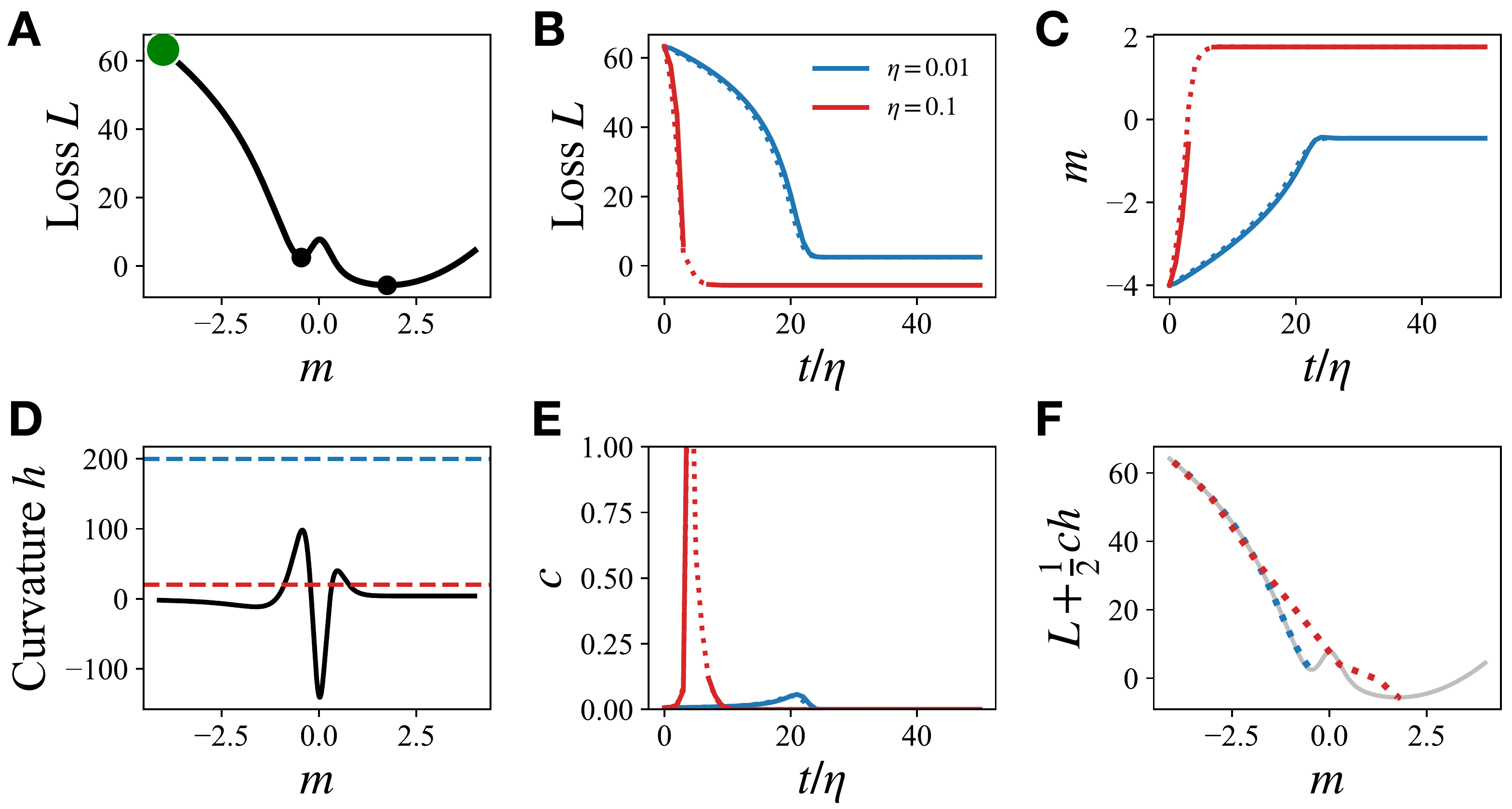}
    \caption{Curvature and parameter covariance determine effective loss landscape.
    \textbf{(A)} A one-dimensional loss landscape with initial parameter value $m(0) = -4$.
    \textbf{(B)} Loss converges to a local or global minimum under gradient descent (dotted lines: simulations; solid lines: theory) with two different learning rates.
    \textbf{(C)} Parameters converge to a local or global minimum.
    \textbf{(D)} Curvature of the loss function; dashed lines are critical values $h_\mathrm{crit} = \tfrac{2}{\eta}$ for each of the learning rates.
    \textbf{(E)} Covariance of the parameter ensemble under gradient descent, given initial variance $c(0) = 0.08^2$.
    \textbf{(F)} Effective loss $L + \tfrac12 ch$ along the trajectory of the ensemble mean (dotted), compared with $L$ (gray).
    }
    \label{fig:1d_gradient_descent}
\end{figure}

These dynamics are illustrated in a one-dimensional example in Figure \ref{fig:1d_gradient_descent}.
In this one-dimensional loss landscape for an ensemble of one-parameter networks with nearly identical initialization, gradient descent with a small learning rate causes the parameter to converge to a sharp local minimum, while larger learning rates enable the parameter to escape the local minimum and converge to a broader, globally optimal solution (Figure \ref{fig:1d_gradient_descent}A-C).
By relating the learning dynamics to the curvature of the loss landscape, the SLF equations characterize the mechanism for this escape. 
Because the curvature at the local minimum exceeds the critical value $2 / \eta$ when $\eta$ is sufficiently large (Figure \ref{fig:1d_gradient_descent}D), the parameter covariance diverges upon approach to the local minimum, providing a mechanism for escape (Figure \ref{fig:1d_gradient_descent}E) (whereupon the SLF solution, which diverges as the ensemble escapes, is restarted from the simulated moments---Methods).
This escape corresponds to a smoothing out of the effective loss function, which includes the implicit regularization term that is dynamically activated in regions of high curvature (Figure \ref{fig:1d_gradient_descent}F).
Together, this simple example illustrates that, even in the absence of batch noise, the finite-step-size corrections captured by SLF provide a mechanism of escape from sharp local minima during learning.

The effects of batch noise, \textit{i.e.}~the random draws of data in each batch, are captured by the terms in the second line of \eqref{eq:20260501b}, as well as by fluctuations in the term $\eta \langle \bH \bC \bH \rangle$.
To begin, consider the term $\mathrm{Cov}(\mathbf{f})$, which characterizes the variance of the gradients computed from different batches of data.
Previous work has shown that, early in training, this covariance is relatively isotropic, with gradients from different batches pushing $\bC$ in different directions \cite{gur2018gradient}.
Later in training, if the loss is a negative log-likelihood and the model approximately matches the data distribution, the Fisher identity (Appendix A) implies $\mathrm{Cov}(\mathbf{f}) \approx \langle H\rangle/B$ \cite{gur2018gradient,zhu2019anisotropic,thomas2020interplay}. This condition does not always hold, however. With noiseless targets, for example, $\mathrm{Cov}(\mathbf{f})$ vanishes at an interpolating minimum. For this reason we evaluate $\mathrm{Cov}(\mathbf{f})$ directly in all numerical experiments.
Unlike the curvature term described above, this term always has the effect of pushing the parameters out of regions of large positive curvature.
The appearance of the coefficient $\eta / B$ is consistent with previous work that have found this ratio to be a key determinant of SGD dynamics, with higher values leading the parameters to wider final minima \cite{jastrzkebski2017three}. 

The last term on the right hand side of \eqref{eq:20260420u}, $-\tfrac{\eta}{2} \left[ 
\mathrm{Cov}(\mathbf{f}, 
\nabla \mathrm{Tr}(\bC \bH))
+ \mathrm{Cov}(\nabla \mathrm{Tr}
(\bC \bH), \mathbf{f})
\right]$,
shows that, in the presence of batch noise and finite step size, higher derivatives of the loss function can influence the parameter dynamics.
To gain insight into this force arising from finite step size, we can consider it along just one dimension:
\begin{equation}
\label{eq:20260426b}
    D_t C \ni -\frac{\eta}{2} C \left[
    \mathrm{Cov}(f, \partial^3 L)
    + \mathrm{Cov}(\partial^3 L, f)
    \right].
\end{equation}
The sign of this term is set by how the batch gradient covaries with the batch third derivative across batches, not by the sign of $\langle \partial^3 L \rangle$ itself, so it can either amplify or damp the covariance. Its counterpart in the mean dynamics \eqref{eq:20260420u} has a definite effect, however. In one dimension, $D_t m \ni -\frac{1}{2} C \langle \partial^3 L \rangle = -\frac{1}{2} C \partial_m \langle h \rangle$, where $h = \partial^2 L$ is the curvature, so the covariance drives the mean toward lower curvature at a rate proportional to $C$. This is the mechanism by which the covariance feeds back on the sharpness, which we analyze in Section \ref{sec:EoS}. 

Together, the dynamical equation for the parameter covariance gives a rich picture of SGD dynamics that is entirely absent from the gradient-flow approximation.
In particular, it describes a tug of war between the $O(\eta^0)$ terms, which push the parameters down into valleys of the loss function; the SGD noise term, which provides exploratory noise early in training followed by a mechanism for escaping local minima later in training; and the remaining $O(\eta)$ terms, in which batch noise provides mechanisms for escape from sharp minima.

\subsection{Stochastic Learning Flow in a two-layer linear network}
\label{sec:DLN}
To test whether the SLF equations capture the noisy dynamics of learning in a setting where the loss landscape is understood analytically, we first applied them to a two-layer deep linear network (DLN) trained with SGD. Although its input--output map is linear, the DLN loss is nonconvex, and its critical points and gradient-flow trajectories are known in closed form \cite{saxe2013exact,saxe2019mathematical,jacot2021saddle}. From small initialization, the network learns the singular modes of the task one at a time, with each mode learned in a rapid transition between a saddle point and the next critical point. 
We used the smallest network that exhibits all of these features: $N_x = N_h = N_y = 2$ units per layer, trained on a task with two distinct singular values, so that the network passes through two saddle points before reaching a minimum. With $P = 8$ parameters, the full Hessian and parameter covariance could be computed and followed mode by mode, and the averages over inputs in \eqref{eq:20260501a}-\eqref{eq:20260501b} could be evaluated exactly (Methods). The network mapped isotropic inputs $x \sim \mathcal{N}(0, I)$ to outputs $\hat{y} = W^2 W^1 x$, which were trained to match targets $y = \Sigma^{yx} x$ by minimizing the mean squared error $\langle L \rangle = \frac{1}{2} \int dx \, p(x) |y(x) - \hat{y}|^2$.

We computed the parameter mean $\bm(t)$ and covariance $\bC(t)$ from the SLF equations \eqref{eq:20260501a}-\eqref{eq:20260501b} as well as $\bm(t)$ in the $\eta \to 0$ gradient-flow (GF) limit.
The results were compared against an ensemble of identically initialized simulated networks trained with SGD all starting from the same initial condition, where stochastic noise during learning led to a distribution of parameters $\bth$ across this ensemble of $N_s=800$ networks, characterized by the empirical mean $\mathbb{E}[\bth(t)] = \tfrac{1}{N_s} \sum_{n=1}^{N_s} \bth^{(n)}(t)$ and covariance $\mathrm{Cov} (\bth(t))$.

\begin{figure}
    \centering
    \includegraphics[width=1.0\linewidth]{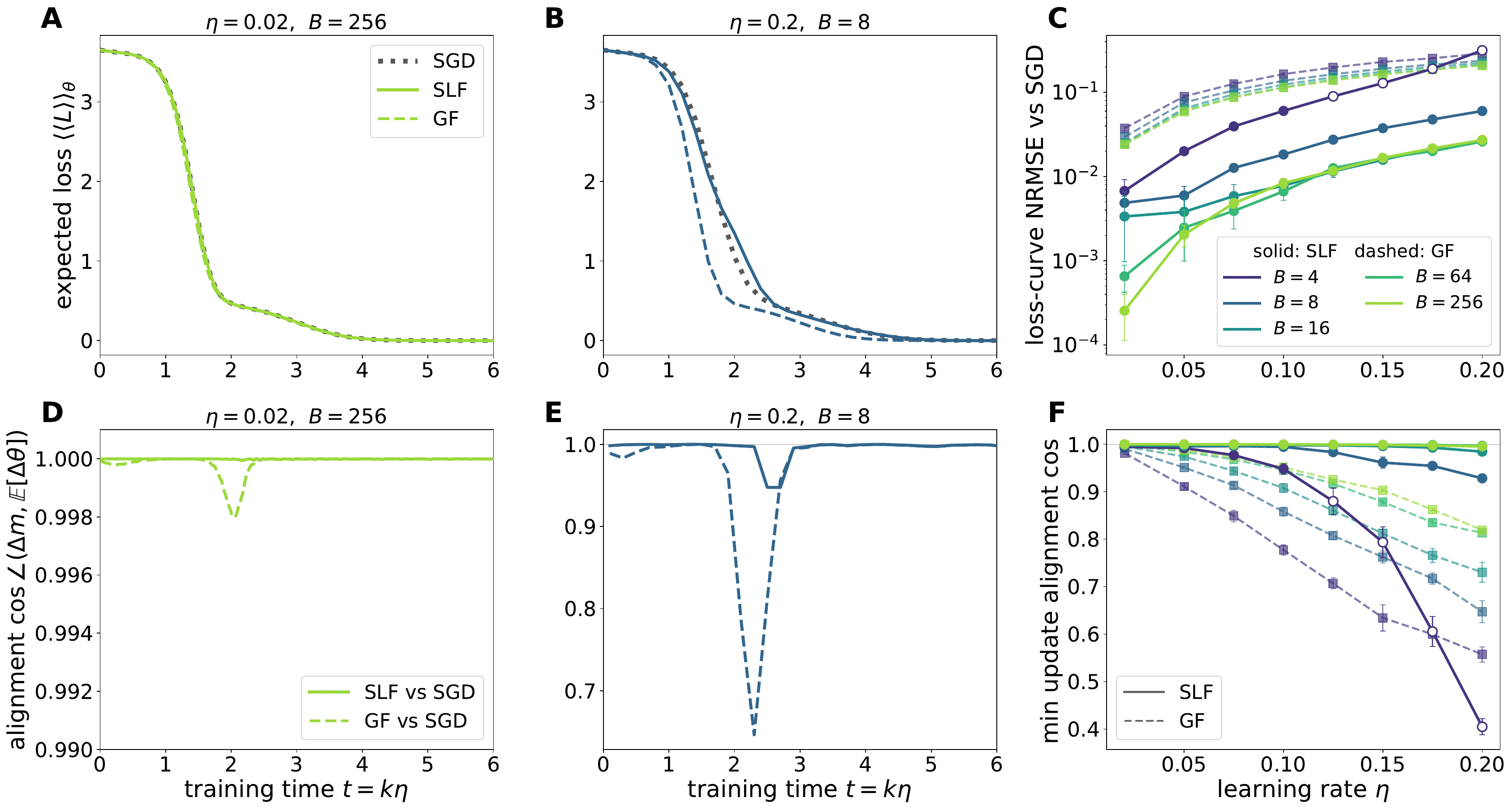}
    \caption{Stochastic Learning Flow captures learning curves and parameter changes in a two-layer deep linear network.
    \textbf{(A)-(B)} Expected loss during training of an ensemble of networks trained with SGD from a single initial condition, as well as corresponding gradient-flow (GF) and Stochastic Learning Flow (SLF) curves in the low-noise (A) and high-noise (B) regimes.
    \textbf{(C)} The normalized root mean square error (NRMSE) of the SLF vs.~SGD and GF vs.~SGD loss curves.
    \textbf{(D)-(E)} Alignment of the ensemble-averaged SGD parameter update direction with the update direction predicted by either GF or SLF in the low-noise (D) and high-noise (E) regimes.
    \textbf{(F)} The minimum update alignment values during training for various learning rates and batch sizes.
    Points in (C) and (F) are mean $\pm$ std across $n=5$ identically initialized networks.
    Open circles denote $(\eta, B)$ settings for which $p(\theta, t)$ is strongly non-Gaussian, limiting the applicability of the SLF theory (\textit{cf.}~Figure \ref{fig:dln-covariance}D).
    }
    \label{fig:dln-loss-alignment}
\end{figure}

While the expected SGD loss curve was described well by either SLF or GF in DLNs trained with small $\eta$ and/or large batch size $B$, SLF captured the SGD learning curve much better than GF as $\eta$ was increased and/or $B$ was decreased (Figure \ref{fig:dln-loss-alignment}A-C).
In addition, while the parameter update direction $\tfrac{d\bm(t)}{dt}$ under either SLF or GF was well aligned with the SGD update direction $\mathbb{E}[\Delta \bth(t)]$ for small $\eta$ and/or large $B$, only SLF was able to capture the SGD parameter updates throughout learning in the noisy learning regime (Figure \ref{fig:dln-loss-alignment}D-F).
Together, these results show that SLF captures the loss curves and mean parameter dynamics of SGD much more faithfully than GF, particularly in the high-noise regime of large $\eta$ and small $B$.

\begin{figure}
    \centering
    \includegraphics[width=1.0\linewidth]{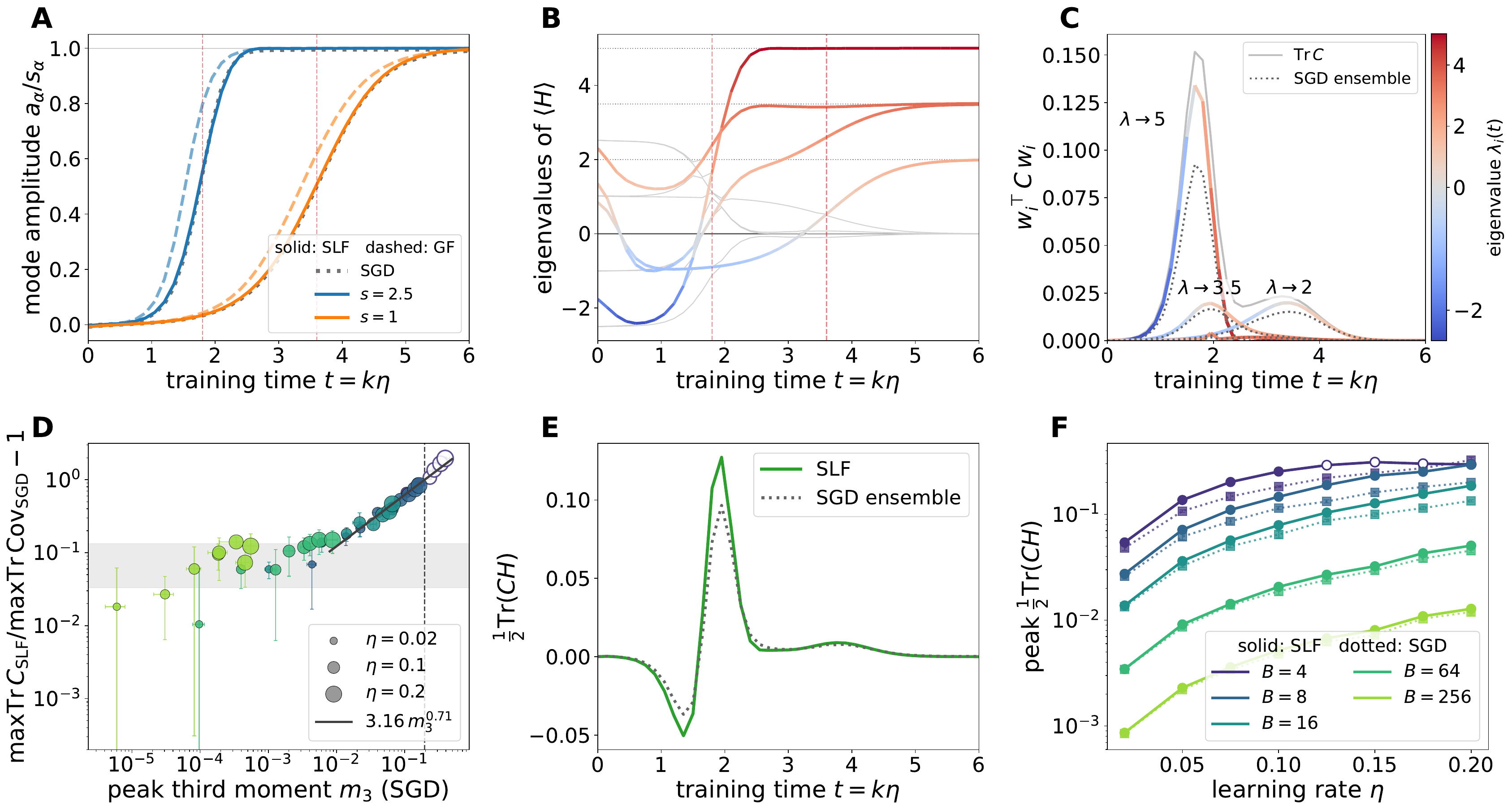}
    \caption{Curvature and covariance dynamics in a deep linear network.
    \textbf{(A)} An ensemble of identically initialized two-layer DLNs learn two task modes with SGD (dotted); learning dynamics are characterized by SLF (solid) and GF (dashed), with vertical lines denoting times of mode learning.
    \textbf{(B)} Eigenvalues of the Hessian during training (colored: output-potent; gray: output-null).
    \textbf{(C)} Parameter covariance from the SGD ensemble projected onto the Hessian eigenmodes (dotted), together with the SLF covariance (colored: mode-projected; gray: total).
    \textbf{(D)} Relative overshoot in the SLF estimate of the covariance peak in (C) increases with the non-Gaussian third moment of the parameter distribution as batch size decreases (colors as in panel F).
    \textbf{(E)} The implicit regularization term from the effective loss function during training.
    \textbf{(F)} The peak values of the implicit regularization during training.
    Points in (D,F) are means $\pm$ std across 5 identically initialized networks.
    Open circles denote $(\eta, B)$ settings for which $p(\theta, t)$ is strongly non-Gaussian, limiting the applicability of the SLF theory.
    }
    \label{fig:dln-covariance}
\end{figure}

To understand the stochastic mechanisms responsible for these learning dynamics, we next investigated the behavior of the parameter covariance matrix $\bC(t)$ as the DLN learned each of the two singular modes of the task matrix $\mathrm{Cov}(\mathbf{y}, \bx)$ (Figure \ref{fig:dln-covariance}A). 
As in previous work, we found that learning each of these modes corresponded with a change in sign of one or more Hessian eigenvalues from negative to positive, indicative of transitions through a sequence of saddle points and ultimately to a minimum \cite{saxe2013exact,saxe2019mathematical,jacot2021saddle,pesme2023saddle,zhang2026saddle} (Figure \ref{fig:dln-covariance}B).
Projecting the parameter covariance along these Hessian modes revealed a sequence of transient expansions along negative-curvature directions, followed by contractions along these directions as the curvature changed sign (Figure \ref{fig:dln-covariance}C).
While the timings of these transient increases precisely matched those in an ensemble of SGD-trained networks, we found that the peak covariance amplitudes from the SLF equations exceeded the SGD ensemble covariance, and that this overshoot grew with the third moment of the SGD ensemble, which itself increased as batch size was reduced (Figure \ref{fig:dln-covariance}C-D).
Given $\bC(t)$ and the time-varying Hessian, we were able to construct the implicit loss regularization from \eqref{eq:20260501c}, which exhibited transient peaks as the network passed through each saddle point (Figure \ref{fig:dln-covariance}E). 
Across a range of learning rates and batch sizes, the dynamics of the implicit regularization term in the integrated flow equations quantitatively matched the SGD ensemble (Figure \ref{fig:dln-covariance}F).
Comparing Figures \ref{fig:dln-covariance} and \ref{fig:dln-loss-alignment} shows that, in the noisy regime of large $\eta$ and small $B$, the magnitude of the regularization term comes within an order of magnitude of the MSE loss.
Together, these results show that the dynamics of the parameter covariance can be predicted by the SLF theory for small and intermediate levels of batch noise and learning rates, and that is accompanied by growing non-Gaussianity of the parameter distribution.

\subsection{Stochastic Learning Flow in a nonlinear network}
\label{sec:nonlinear}
To test the ability of the SLF equations to characterize learning dynamics in a nonlinear network, where less is known analytically about the loss landscape and learning dynamics, we next trained a network to perform a binary classification task.
Because validation of the theory requires a large number of replicas to be trained for each setting of parameters and hyperparameters, we implemented a small-scale two-layer classifier on two digits (0 \textit{vs.} 1) from the MNIST dataset.
Each test began from a common parameter vector reached partway through training, so that $\bC(0) = 0$ and the covariance was generated entirely by minibatch noise, and followed the ensemble for 200 steps, beyond which the Gaussian approximation began to break down at intermediate batch sizes (Methods).

We first asked whether including the covariance in the parameter flow equations increased the ability of the model to predict the evolution of the parameters over time.
\begin{figure}
    \centering
    \includegraphics[width=0.8\linewidth]{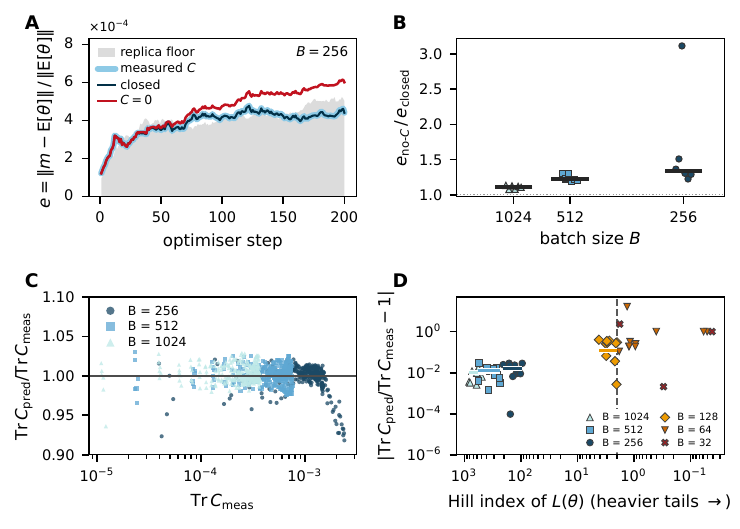}
    \caption{Stochastic Learning Flow in a nonlinear network.
    \textbf{(A)} In a two-layer network trained on binary MNIST, the integrated SLF equations (black) predict the mean parameter trajectory of an ensemble of $n=256$ SGD-trained replicas to within estimation noise (gray). The prediction is unchanged if SLF uses the replica covariance rather than the integrated covariance (blue), but degrades if the parameter covariance is neglected in the flow equations (red).
    \textbf{(B)} Ratio of the parameter error at step $k=200$ assuming $\bC=0$ (\textit{cf}.~red line in (A)) to the error of the full SLF equations (\textit{cf}.~black line in (A)) (6 initial conditions at each batch size; bars are medians).
    \textbf{(C)} Ratio of predicted covariance trace to measured covariance trace as a function of the measured covariance trace, which increases with decreasing batch size (for each $B$, points correspond to optimizer steps and network initializations).
    \textbf{(D)} Relative covariance error \textit{vs}.~Hill index of the distribution of losses across replicas, quantifying heaviness of the distribution tail (points are 10 timesteps over $0<k\leq50$ from a single network for each $B$; dashed line: index 2, where the 2nd moment of the distribution diverges; colored lines: median sampling precision of $\mathrm{Tr}\,\bC_\mathrm{meas}$ in cases where they can be estimated reliably). 
    }
    \label{fig:nonlinear}
\end{figure}
The error during training was computed by comparing the mean parameter vector $\bm$ from the integrated SLF equations to the ensemble mean from a set of SGD-trained networks with identical initializations (Figure \ref{fig:nonlinear}A).
Integrating the coupled SLF equations \eqref{eq:20260501a}-\eqref{eq:20260501b} yielded parameters that were within the estimation noise floor from the finite SGD ensemble (Figure \ref{fig:nonlinear}A, black line).
Further, computing $\bm(t)$ from \eqref{eq:20260501a} using the measured $\bC(t)$ from the SGD ensemble rather than the integrated $\bC(t)$ from \eqref{eq:20260501b} led to equally good prediction (Figure \ref{fig:nonlinear}A, blue line).
In contrast, setting $\bC(t)=0$ in the flow equation impaired the prediction (Figure \ref{fig:nonlinear}A, red line), with the size of this effect growing with decreasing batch size (Figure \ref{fig:nonlinear}B), showing that, in the regimes we tested, $\bC(t)$ arising from batch noise plays a small but significant role in determining the parameter evolution.

We next quantified the extent to which the parameter covariance $\bC(t)$ predicted by the SLF equations matched the empirical parameter covariance measured across the SGD ensemble.
Because of the high dimensionality of the parameter space, which makes comparing covariance matrices using a finite number of observations difficult, we quantified the similarity of the overall variance by computing the traces of the covariance matrices.
At low and intermediate levels of batch noise, we found that the SLF equations successfully predicted the overall variance to within $10\%$ across two orders of magnitude (Figure \ref{fig:nonlinear}C).

We hypothesized that increased error in the variance prediction relative to measured values at higher levels of batch noise might be, as in the DLN studied in Section \ref{sec:DLN}, associated with non-Gaussian features of the parameter distribution emerging through training.
Because of the higher dimensionality of the parameter space in this nonlinear network, higher moments of the parameter distribution itself could not be accurately computed from the finite number of SGD replicates used in our simulation.
Hence, we considered the one-dimensional distribution of losses over the first 50 steps of training and used the Hill index to quantify the heaviness of the distribution's tail, where an index value of $<2$ indicates divergence of the distribution's second moment.
Heavy tails are expected because multiplication by random batch Hessians makes SGD a random-coefficient recursion, whose stationary distributions are generically power-law \cite{kesten1973random,gurbuzbalaban2021heavy,hodgkinson2021multiplicative}.
The error of the predicted variance grew as the Hill index fell (\ref{fig:nonlinear}D). 
In cases where the 2nd moment existed and the sampling precision of $\mathrm{Tr}\,\bC$ could be reliably estimated, we found that this sampling precision closely tracked the variance prediction error, indicating that the Gaussian closure and the variance estimator fail together because both are second-moment objects, and consistent with both being driven by heavy tails.
   
Together, these results show that there exists an intermediate regime of batch noise in which SLF successfully predicts parameters and their covariances during training, and that the breakdown of these predictions as batch noise increases further is associated with non-Gaussian effects.

\subsection{Edge of Stability as a dynamical stationary point}
\label{sec:EoS}
Previous work \cite{cohen2021gradient} found empirically that deep networks trained with full-batch gradient descent operate at the \textit{Edge of Stability} (EoS), with the maximum eigenvalue of the loss Hessian rising under training and then hovering near $2/\eta$. 
Two opposing forces contribute to this phenomenon. 
The first, progressive sharpening, is the tendency of loss curvature to increase under gradient descent, pushing the solution toward the threshold from below \cite{cohen2021gradient,thomas2020interplay}. 
The second arises from finite step size and forces the solution out of regions of high curvature \cite{lewkowycz2020large,cohen2021gradient,damian2023self}, as seen in Figure \ref{fig:1d_gradient_descent}D-E. 
In this section, we show that the EoS emerges as a stationary point of the coupled dynamics \eqref{eq:20260501a}-\eqref{eq:20260501b} in the full-batch limit, and that it is shifted downward by noise arising from finite batch size.

To begin, consider full-batch gradient descent and project along the direction $\bv_1$ of sharpest curvature, for which $\bH \bv_1 = \lambda_1 \bv_1$, where $\lambda_1$ is the largest eigenvalue of $\bH$.
Let $\sigma_1^2 = \bv_1^\top \bC \bv_1$ be the covariance, where covariance along other directions is suppressed by \eqref{eq:20260426a}.
Then
\begin{equation}
\label{eq:20260805a}
\frac{d \sigma_1^2}{dt} = \bv_1^\top \bC \bv_1,
\end{equation}
where, because $\bv_1$ is normalized, we used $\tfrac{d}{dt}(\bv_1^\top \bv_1) = 0$.
In this case, \eqref{eq:20260501b} gives
\begin{equation}
\label{eq:20260806b}
D_t \sigma_1^2 = \lambda_1 (\eta \lambda_1 - 2) \sigma_1^2.
\end{equation}
This equation shows that $\lambda_1 = \tfrac{2}{\eta}$ defines the EoS, with $\sigma_1^2(t)$ increasing (decreasing) for sharpness above (below) this value.
In particular, the existence of this critical value depends on the finite-step-size term $\eta \lambda_1^2$, which arises from the term $\eta \langle \bH \rangle \bC \langle \bH \rangle$ in the covariance flow and is absent in first-order SDE approaches.
Coupled to these dynamics are the dynamics of $\lambda_1$ itself, which evolves during learning as $\partial_t \lambda_1 = (\nabla \lambda_1)^\top \dot{\bm}$. 
Using the fact that $\dot{\bm} = \bF + O(\eta)$, where $\bF = -\nabla \langle L \rangle - \tfrac12 \nabla \mathrm{Tr}(\bC \langle \bH \rangle)$, this gives
\begin{equation}
\label{eq:20260807a}
D_t \lambda_1 = (\nabla \lambda_1)^\top D_t \bm
+ \frac{\eta}{2} (D_t \bm)^\top 
(\nabla \nabla \lambda_1) D_t \bm.
\end{equation}
Dropping the terms of $O(\eta^2)$, this is
\begin{equation}
\label{eq:20260807b}
D_t \lambda_1 
= S_1 - \frac12 g_1 \sigma_1^2 + \frac{\eta}{2} Q_1,
\end{equation}
where we have defined
\begin{align}
\begin{split}
\label{eq:20260807c}
S_1 &= - (\nabla \lambda_1)^\top 
(\nabla \langle L \rangle), \\
g_1 &= |\nabla \lambda_1|^2, \\
Q_1 &= \bF^\top (\nabla \nabla \lambda_1) \bF.
\end{split}
\end{align}
Here $S_1$ is the rate of progressive sharpening under gradient flow, which previous empirical and theoretical work has demonstrated tends to be positive in multilayer networks \cite{cohen2021gradient,thomas2020interplay}.
The second term in \eqref{eq:20260807b} shows that covariance back-reaction (which, in full-batch learning, corresponds to period-2 oscillations) tends to counteract progressive sharpening, providing the mechanism for escape from sharp minima; this is the self-stabilization mechanism of Ref.~\cite{damian2023self}, expressed through the covariance.
The third term is the curvature of the sharpness landscape along the direction of parameter drift and arises due to finite step size.

The dynamical equations \eqref{eq:20260806b} and \eqref{eq:20260807b} exhibit a stationary point at
\begin{equation}
\label{eq:20260807d}
\lambda_1^* = \frac{2}{\eta}, \quad
\sigma_1^{2*} = \frac{2S_1 + \eta Q_1}{g_1}.
\end{equation}
In full-batch gradient descent, $\sigma_1^2$ is the variance of an ensemble of trajectories or, for a single trajectory, the squared amplitude of its period-2 oscillation about the two-step average $\frac{1}{2}(\theta_{k-1} + \theta_k)$. Because $\sigma_1^2 = 0$ is unstable for $\lambda_1 > 2/\eta$ by (29), any initial spread, however small, grows until it reaches $\sigma_1^{2*}$.
Equation \eqref{eq:20260807d} shows that the EoS threshold corresponds to the unique stationary state of the covariance dynamics.
Further, we note that gradient flow corresponds to the $\sigma_1^2 = 0$ limit of this theory, in which case $\sigma_1^2$ remains 0 by \eqref{eq:20260806b}, and the curvature evolves according to $D_t \lambda_1 = S_1 + \tfrac{\eta}{2}Q_1$, which possesses no mechanism to arrest progressive sharpening.

For $\eta Q_1\to0$ this is central flow \cite{cohen2025understanding}, in which the mean descends on $L+\frac12\sigma_1^2\lambda_1$, the effective loss \eqref{eq:20260501c} with $\mathbf C=\sigma_1^2v_1v_1^{\mathsf T}$, and $\sigma_1^2=2S_1/g_1$ is fixed by imposing $\dot\lambda_1=0$ at $\lambda_1=2/\eta$. Here the same values follow from the stationary state of the covariance dynamics rather than being imposed, and \eqref{eq:20260806b}-\eqref{eq:20260807a} also describe the approach to the edge. Central flow corresponds to the limit in which the relaxation of $\sigma_1^2$ is replaced by a constraint.

In the case where the batch size $B$ is finite, the dynamical equations for the parameter covariance and loss curvature contain corrections due to the effects of batch noise.
The same projection that was used above in this case gives
\begin{equation}
\label{eq:20260807e}
D_t \sigma_1^2 = 
-\rho(\lambda_1) \sigma_1^2 + \frac{\eta c}{B},
\end{equation}
where
\begin{equation}
\label{eq:20260807f}
\rho(\lambda_1) 
= -\eta \left( 1 + \frac{\kappa}{B} \right) \lambda_1^2
+ 2 \lambda_1
- \frac{\eta}{B}d.
\end{equation}
Here we have defined
\begin{align}
\label{eq:20260807g}
\begin{split}
c &= \mathrm{Var}_i (\bv_1^\top \nabla \ell_i), \\
d &= \mathrm{Cov}_i (\bv_1^\top \nabla \ell_i,
\bv_1^\top \nabla h_i),
\end{split}
\end{align}
where $h_b = \tfrac{1}{B} \sum_{i \in b} h_i$ is the batch curvature, and the sample curvature $h_i$ satisfies $\mathbb{E}_i[h_i] = \lambda_1$. 
Letting $\mathrm{Var}_i[h_i] = \kappa \lambda_1^2$, the statistics of $h_b$ are then defined by $\mathbb{E}_b[h_b] = \lambda_1$ and $\mathrm{Var}_b[h_b] = \tfrac{\kappa \lambda_1^2}{B}$.
Here $\mathbb{E}_i$, $\mathrm{Var}_i$ and $\mathrm{Cov}_i$ denote averages over the $N$ training samples, and $\mathbb{E}_b$ and $\mathrm{Var}_b$ averages over batches, so that $\mathbb{E}_b[\cdot] = \langle \cdot \rangle$.

The variance along $\bv_1$ remains finite only while $\rho > 0$, so batch noise lowers the threshold at which it diverges to the point where $\rho(\lambda_0) = 0$, which to first order in $d$ is
\begin{equation}
\label{eq:20260807j}
\lambda_0 = \frac{2}{\eta \left( 1 + \frac{\kappa}{B} \right)}
- \frac{\eta d}{2B},
\end{equation}
consistent with the observation that SGD hovers below $\tfrac{2}{\eta}$ \cite{cohen2021gradient,andreyev2024edge}.
The first term in \eqref{eq:20260807j} is the mean-square linear stability threshold of SGD projected onto a single direction. That batch-to-batch curvature fluctuations lower this threshold was identified in Ref.~\cite{wu2018sgd}, and the exact mean-square condition was derived (assuming batches sampled without replacement) in Ref.~\cite{mulayoff2024exact}.
At each step, the parameter variance along $\bv_1$ is multiplied by $(1 - \eta h_b)^2$, which is convex in $h_b$, so that fluctuations in $h_b$ cause marginal stability to arrive earlier.
The second term arises from the coupling of gradient noise to the third derivative of the loss; it vanishes at interpolating minima, where every per-sample gradient is zero, which is why it is absent from the linear stability analysis of Ref.~\cite{mulayoff2024exact}.
Supplemental Figure \ref{fig:eos} shows that, while \eqref{eq:20260807j} successfully captures the reduction in $\lambda_0$ due to batch noise in a nonlinear network using a probe test, versions of \eqref{eq:20260807j} with \textit{vs.}~without the second term were identical to within $<1\%$, indicating that the $d$ term plays a comparatively small role in determining the EoS in the network we tested.

Two limitations apply to the result in \eqref{eq:20260807j}. 
First, $\lambda_0$ is a mean-square threshold---beyond it the variance diverges, while typical trajectories can remain bounded up to a Lyapunov threshold lying, to leading order, as far above $2/\eta$ as $\lambda_0$ lies below it \cite{kesten1973random,ziyin2023type}.
Second, because the back-reaction in \eqref{eq:20260807a} is the rank-one contraction of $(\nabla\lambda_1)^{\mathsf T}\nabla\mathrm{Tr}(\mathbf C\langle\mathbf H\rangle)$, which depends on the whole covariance once batch noise populates its other modes, the noise amplitude in \eqref{eq:20260807d} does not carry over to finite $B$. 


Together, the results in this section show that the EoS can be understood as a dynamical stationary point of the SLF equations in the full-batch limit, and that the EoS stability threshold is shifted downward by noise arising from finite batch size.

\section{Discussion}
We have derived a set of coupled dynamical equations describing the evolution of the first two moments of the parameter distribution over an ensemble of networks, where these moments include noise effects arising from random sampling of inputs and finite step size.
Applying these equations to SGD, we found that the theory describes gradient flow on an effective loss function that couples the parameter mean and covariance, the evolution of which depends on the loss curvature together with noise effects arising from finite step size and batch noise.
Applying the flow equations to deep linear and nonlinear networks, we found that they enable successful prediction of loss curves, parameter updates, and parameter covariance at low and intermediate levels of noise, and we identified the regime in which the predictions begin to break down as non-Gaussian features of the parameter distribution become more pronounced.
Finally, in the full-batch limit of gradient descent, the theory recovers the edge of stability as a stationary point of a pair of coupled equations describing the dynamics along the sharpest direction in the loss landscape, while, for finite batch size, the theory predicts the lowering of the EoS stability threshold as batch noise increases.

An important limitation of the SLF theory is the truncation of the moment equations at second order.
As shown in Figures \ref{fig:dln-covariance} and \ref{fig:nonlinear}, the emergence of non-Gaussian features in the parameter and loss distributions coincides with a decrease in the ability of SLF to predict the dynamics of the distribution.
While addressing this by extending the approach of \eqref{eq:20260420j} and \eqref{eq:20260420l} to include higher moments would be possible in principle and could extend the ability of the theory to capture dynamics as the distribution becomes non-Gaussian, the complexity of the resulting equations would make interpretation more difficult, and the theory would still be unable to capture the possible fragmentation of the parameter distribution into multiple modes as replicas settle into different minima later in training.
Future work could address this by treating the few directions along which non-Gaussian structure develops separately, for example with a reduced Fokker–Planck description or a description of the tails in the spirit of random-coefficient recursions~\cite{kesten1973random}, while retaining the second-moment description elsewhere.
For the EoS theory, an additional limitation is the projection of the parameter dynamics along a single direction.
While this is justified in the case of full-batch gradient descent, our results for the case of SGD show that the sharp mode becomes coupled to other modes due to the injection of variance along additional directions arising from batch noise.
Developing a dynamical theory to fully capture EoS phenomena under SGD by including this mode coupling remains an important direction for future work.
Finally, due to the high computational cost of solving the flow equations, which requires $P\times P$ covariance matrices and dense Hessian computations, the networks tested here have at most a few thousand parameters, so whether the regime of validity identified in Sections \ref{sec:DLN}-\ref{sec:nonlinear} persists in networks with more than a few thousand parameters remains to be determined.

A widely used approach models SGD by an SDE whose drift is the full-batch gradient and whose diffusion is the minibatch gradient covariance~\cite{mandt2016variational,jastrzkebski2017three,chaudhari2018stochastic,ali2020implicit,chen2023stochastic}.
Several of these analyses fix the diffusion to be constant near an optimum~\cite{mandt2016variational} or proportional to the Hessian~\cite{jastrzkebski2017three}, which removes the batch-curvature fluctuation and cross terms of \eqref{eq:20260501b}.
Other approaches retain its dependence on $\bth$~\cite{chaudhari2018stochastic,ali2020implicit,chen2023stochastic}, which generates these terms.
Li \textit{et al.}~\cite{li2017stochastic} showed that this SDE is a first-order weak approximation of SGD and derived a second-order one in which the drift descends on $L+\tfrac{\eta}{4}|\nabla L|^2$, which was also derived by backward error analysis and analyzed numerically in Ref.~\cite{barrett2020implicit} and which can be obtained in SLF by eliminating the $\partial_t^2$ term of \eqref{eq:20260501a} perturbatively at $\mathbf C=0$.
These approximations have since been refined \cite{gess2023stochastic}.

What all of these SDE descriptions lack is the deterministic finite-step part of the covariance dynamics: the term $\eta\langle\bH\rangle\bC\langle\bH\rangle$ and the operator $D_t$. The difference matters where $\eta\lambda$ is not small. The second-order SDE expands the one-step amplification of a perturbation in powers of $\eta\lambda$, a series that converges only for $\eta\lambda<1$, whereas the SLF covariance equation retains the one-step change of the second moment, which for a quadratic loss is exact in $\eta\lambda$. 
Accordingly, near a quadratic minimum the stationary solution of \eqref{eq:20260501b} coincides with the exact stationary covariance of discrete-time SGD \cite{liu2021noise}, including its divergence at $\eta\lambda=2$. Moment relations obtained directly from the discrete update have also been used to derive fluctuation–dissipation relations at stationarity \cite{yaida2019fluctuation}; SLF extends this approach to the full time-dependent dynamics.
This is why the SLF equations contain the edge-of-stability threshold, and may explain why SDE surrogates of SGD, even simulated at finite step size, do not stabilize where SGD does~\cite{andreyev2024edge}. Kunin \textit{et al.}~\cite{kunin2023limiting} describe the late phase of SGD with an underdamped equation whose inertial term, for plain SGD, coincides with that of
$D_t$. Because they apply it to individual trajectories rather than to the parameter distribution,
however, their covariance dynamics lack $\eta\langle\bH\rangle\bC\langle\bH\rangle$ and the
instability at $\eta\lambda=2$.
The SLF equations bring together several existing descriptions of the edge of stability. The back-reaction of the covariance on the sharpness is the self-stabilization mechanism of Damian et al.~\cite{damian2023self}, expressed through the covariance; central flow~\cite{cohen2025understanding} is the full-batch limit in which the relaxation of the covariance is replaced by a constraint; and the batch-size shift of the threshold coincides with the mean-square linear stability analysis of SGD~\cite{mulayoff2024exact}, with an additional term from the coupling of gradient noise to the third derivative of the loss. Liao \textit{et al.}~\cite{liao2026sgd} extended self-stabilization to SGD and derived a closed-form displacement of the sharpness below $2/\eta$ from additive gradient noise, with the threshold held at $2/\eta$. In our framework that displacement depends on the whole covariance once batch noise populates modes other than the sharpest, so a description in terms of the sharp direction alone is incomplete at finite batch size. Two further questions remain open: where between the mean-square and Lyapunov thresholds trajectories settle~\cite{ziyin2023type}, and how additional modes are recruited to the edge, as observed empirically~\cite{cohen2021gradient}.

While we have analyzed SLF for the particular case of SGD, the general flow equations \eqref{eq:20260420r}-\eqref{eq:20260420s} apply to any stochastic update rule. Characterizing how the flow of the parameter distribution depends on the learning rule is a potentially fruitful direction for future work. This includes momentum and adaptive methods, in which momentum is known to change the inertial term~\cite{kunin2023limiting}, and reinforcement learning, in which exploratory noise adds to sampling noise. It also includes regularization terms added to the loss, and learning-rate schedules, for which the identification $\delta t=\eta$ used here would have to be generalized. In addition, future work can use the theory to characterize the evolution of hidden-layer representations in multilayer networks, and how these learned representations depend on SGD noise, the learning rule and the loss function. The implicit regularization term in \eqref{eq:20260501c} also connects the covariance dynamics to the long-standing question of how SGD noise biases training toward flat regions of the loss landscape, which the theory could help address quantitatively. Together, the framework presented here treats the parameter covariance as a dynamical variable on the same footing as the mean, and so places the effects of finite step size and sampling noise within reach of the tools of dynamical systems theory that have been used to analyze gradient flow.

\pagebreak

\printbibliography

\pagebreak

\section*{Appendix A: Geometry of the gradient noise}
\label{app:gradient_noise}
To see the relationship between the SGD noise term $\mathrm{Cov}(f_k, f_l)$ in \eqref{eq:20260420u} and the Hessian, we can make use of the Fisher identity \cite{lehmann1998theory}:
\begin{equation}
\label{eq:20260423a}
\mathrm{Cov}_{p(y | x, \theta)}(\partial_k \ell, \partial_l \ell) 
    = \mathbb{E}_{p(y | x, \theta)}[H_{kl}],
\end{equation}
where we have assumed that the model loss can be expressed as a negative log likelihood $\ell (x) = -\ln p(y | x, \theta)$. 
To see this, we write
\begin{equation}
\begin{split}
\label{eq:20260423b}
    0 &= \partial_l \partial_k(1) \\
    &= \partial_l \partial_k \int dy ~p(y | x, \theta) \\
    &= \partial_l \int dy ~p(y | x, \theta) \partial_k \ln p(y | x, \theta) \\
    &= \int dy ~p(y | x, \theta) \left[
    (\partial_k \ln p(y | x, \theta)) (\partial_l \ln p(y | x, \theta))
    + \partial_k \partial_l \ln p(y | x, \theta),
    \right]
\end{split}
\end{equation}
which, using $\ell (x) = -\ln p(y | x, \theta)$, is exactly the Fisher identity \eqref{eq:20260423a}.

Next, if we further assume that the model is close to convergence, such that $p_\mathrm{data}(x,y) \approx p_\mathrm{model}(x,y)$, where $p_\mathrm{model}(x,y) = p(y|x, \theta) p_\mathrm{data}(x)$, we have $\mathbb{E}_{x,y \sim p_\mathrm{data}(x,y)}[\ldots] \approx \mathbb{E}_{x \sim p_\mathrm{data}(x)} [\mathbb{E}_{y \sim p(y|x)}[\ldots]]$.
Applying this to $f_k f_l$ and making use of \eqref{eq:20260423a} then gives 
\begin{equation}
\label{eq:20260423c}
\mathbb{E}_{x,y \sim p_\mathrm{data}(x,y)}[f_k f_l]
\approx \mathbb{E}_{x,y \sim p_\mathrm{data}(x,y)}[H_{kl}].
\end{equation}
Finally, given the definition $f_k = \tfrac{1}{B} \sum_{b=1}^B \partial_k \ell(x^b)$, we have 
\begin{equation}
\begin{split}
\label{eq:20260423d}
    \mathrm{Cov}(f_k, f_l) 
    &= \frac{1}{B} \mathrm{Cov}(\partial_k \ell, \partial_l \ell) \\
    &= \frac{1}{B} \mathbb{E}_{x,y \sim p_\mathrm{data}(x,y)}[f_k f_l] \\
    &\approx \frac{1}{B} \mathbb{E}_{x,y \sim p_\mathrm{data}(x,y)}[H_{kl}].
\end{split}
\end{equation}
This shows that, when the model is close to convergence, the SGD noise is proportional to the Hessian.

\pagebreak

\section*{Methods}

\subsection*{One-dimensional example (Figure 1)}

\paragraph{Loss and ensembles.} The loss was
\begin{equation*}
L(m) = \frac{c_\ell (m + \frac{1}{2})^2 [1 - \tanh(s m)]}{1 + m^2} + c_r (m - \frac{1}{2})^2 [1 + \tanh(s m)] - \gamma m,
\end{equation*}
with $c_\ell = 30$, $c_r = 1$, $s = 3$ and $\gamma = 5$. All derivatives of $L$ were evaluated by central differences with step $2 \times 10^{-4}$. For each $\eta$, an ensemble of $10^4$ replicas was initialized from $\mathcal{N}(-4, 0.08^2)$ and trained by full-batch gradient descent for 50 steps.

\paragraph{SLF equations.} With $\tau = t/\eta$, the one-dimensional, noise-free form of (23)--(24) becomes $\partial_\tau^2 u = 2\eta R_u - 2\partial_\tau u$ for $u \in \{m, c\}$, with $R_u$ the right-hand side. This system was integrated by fourth-order Runge--Kutta with step $\Delta\tau = 5 \times 10^{-4}$, from the ensemble's initial sample mean and variance with zero initial velocities. Integration was halted when $c > 2$, left blank for three steps, and restarted once from the simulated moments and their finite-difference $\tau$-derivatives. At $\eta = 0.1$ this occurred at step 4, with the restart at step 7; at $\eta = 0.01$ it did not occur. Panel F shows $L(\mathrm{E}[\theta]) + \frac{1}{2}\mathrm{Var}(\theta) h(\mathrm{E}[\theta])$ from the simulated ensemble.

\subsection*{Deep linear network (Figures~2 and~3)}

\paragraph{Network and task.} 
The network had $N_x=N_h=N_y=2$ units per layer, $\hat{\mathbf y}=\mathbf W^2\mathbf W^1\mathbf x$, so $\bth=(\mathrm{vec}\,\mathbf W^1,\mathrm{vec}\,\mathbf W^2)$ had $P=8$ components. Inputs were drawn as $\mathbf x\sim\mathcal N(0,\mathbb I)$ and targets were $\mathbf y=\bSigma^{yx}\mathbf x$ with $\bSigma^{yx}=\mathbf U\,\mathrm{diag}(s_1,s_2)\mathbf V^{\top}$, $(s_1,s_2)=(2.5,1)$, and $\mathbf U,\mathbf V$ random orthogonal matrices drawn once with a fixed seed. Training was online, and each step used a fresh batch of $B$ inputs, with batch loss $L=\frac{1}{2}\mathrm{Tr}[(\mathbf W-\bSigma^{yx})\hat{\bSigma}(\mathbf W-\bSigma^{yx})^{\top}]$, where $\mathbf W=\mathbf W^2\mathbf W^1$ and $\hat{\bSigma}=B^{-1}\sum_{n=1}^B\mathbf x_n\mathbf x_n^{\top}$. Every run started from the same balanced initialization, $\mathbf W^1=\sqrt{\epsilon}\,\mathbf O_1^{\top}$ and $\mathbf W^2=\sqrt{\epsilon}\,\mathbf O_2$, with $\epsilon=0.01$ and $\mathbf O_1,\mathbf O_2$ random orthogonal matrices drawn once with a fixed seed. Runs lasted to $t=k\eta=8$.

\paragraph{SGD ensembles.} 
Each ensemble comprised $N_s=800$ replicas started from the initial condition above, each with an independent stream of batches. No replica diverged in any of the runs reported. The empirical mean $\mathbb E[\bth(t)]$ and covariance $\mathrm{Cov}(\bth(t))$ were recorded at every step. The SGD expected loss was the mean over replicas of the population loss $\frac{1}{2}|\mathbf W-\bSigma^{yx}|_F^2$, after discarding the largest $1\%$ of values.

\paragraph{SLF and GF.} 
The SLF equations \eqref{eq:20260501a}-\eqref{eq:20260501b} were integrated with population averages over $\mathbf x$ evaluated in closed form. Because $\mathbf f_b$ and $\mathbf H_b$ are linear in $\hat{\bSigma}$, the moments $\langle\hat\Sigma_{ab}\rangle=\delta_{ab}$ and $\langle\hat\Sigma_{ab}\hat\Sigma_{cd}\rangle=\delta_{ab}\delta_{cd}+B^{-1}(\delta_{ac}\delta_{bd}+\delta_{ad}\delta_{bc})$ give $\langle\mathbf f\rangle$, $\langle\mathbf H\rangle$, $\mathrm{Cov}(\mathbf f)$, $\langle\mathbf H\mathbf C\mathbf H\rangle$, $\nabla\mathrm{Tr}(\mathbf C\langle\mathbf H\rangle)$ and the cross term $\mathrm{Cov}(\mathbf f,\nabla\mathrm{Tr}(\mathbf C\mathbf H))$ exactly. With $D_t=\partial_t+(\eta/2)\partial_t^2$, each equation was written as a first-order system, $\dot{\mathbf m}=\mathbf v$, $\dot{\mathbf v}=(2/\eta)(\mathbf R_m-\mathbf v)$, and likewise for $\mathbf C$, where $\mathbf R_m$ and $\mathbf R_C$ are the right-hand sides of \eqref{eq:20260501a}-\eqref{eq:20260501b}. The system was integrated with an adaptive Runge--Kutta solver (RK45; relative and absolute tolerances $10^{-4}$ and $10^{-6}$) over one interval $[k\eta,(k+1)\eta]$ at a time, from $\mathbf m(0)=\bth(0)$ and $\mathbf C(0)=0$, with initial velocities $\mathbf v(0)=\mathbf R_m$ and $\dot{\mathbf C}(0)=\mathbf R_C$. The loss depends on $(\mathbf W^1,\mathbf W^2)$ only through $\mathbf W$, so it is invariant under $\mathbf W^1\to\mathbf G^{-1}\mathbf W^1$, $\mathbf W^2\to\mathbf W^2\mathbf G$ for any invertible $\mathbf G$. At the end of each interval, $\mathbf C$ and $\dot{\mathbf C}$ were therefore projected onto the $(P-N_h^2)$-dimensional complement of the tangent space of this symmetry at $\mathbf m$. The SLF expected loss was $\langle L(\mathbf m)\rangle+\frac{1}{2}\mathrm{Tr}(\mathbf C\langle\mathbf H\rangle)$. GF solved $\dot{\mathbf m}=\langle\mathbf f\rangle$ with $\mathbf C=0$ (RK45; tolerances $10^{-9}$ and $10^{-11}$). All three descriptions were compared on the grid $t_k=k\eta$.

\paragraph{Figure 2.} 
For each $(\eta,B)$, SLF and GF were run once and the SGD ensemble five times with independent batch streams. The normalized root mean square error was $\mathrm{NRMSE}=\{\overline{(\langle L\rangle_{\rm th}-\langle L\rangle_{\rm SGD})^2}\}^{1/2}/\sigma(\langle L\rangle_{\rm SGD})$, where $\langle L\rangle_{\rm th}$ is the expected loss predicted by SLF or GF, $\langle L\rangle_{\rm SGD}$ is the SGD expected loss, and the bar and $\sigma$ denote the mean and standard deviation over all steps. For the alignment, consecutive updates $\Delta\mathbf m_k=\mathbf m(t_{k+1})-\mathbf m(t_k)$ and $\Delta\mathbb E[\bth]_k$ were projected onto the physical subspace at $\mathbb E[\bth(t_k)]$ before their cosine was taken. Steps at which the SGD update norm was below $10^{-3}$ of its maximum were discarded, and the cosine was averaged over a moving window of $0.05$ time units. Panel F shows the minimum of this smoothed cosine over the run.

\paragraph{Figure 3.} 
Panels A--C and E used $(\eta,B)=(0.15,16)$. Mode amplitudes were $a_\alpha=\mathbf u_\alpha^{\top}\mathbf W\mathbf v_\alpha$, with $\mathbf u_\alpha,\mathbf v_\alpha$ the columns of $\mathbf U,\mathbf V$. Vertical lines mark the times at which the SGD ensemble mean first reached $a_\alpha=s_\alpha/2$. Eigenvalues and eigenvectors $\mathbf w_i$ of $\langle\mathbf H\rangle$ were computed within the physical subspace at each step, and branches were followed through crossings by maximizing the absolute overlap of eigenvectors between consecutive steps (Hungarian assignment). Panel C shows $\mathbf w_i^{\top}\mathbf C\mathbf w_i$ for SLF and, with eigenvectors computed at the ensemble mean, for the gauge-projected $\mathrm{Cov}(\bth)$. In panel E, the SGD value was $\frac{1}{2}\mathrm{Tr}(\mathbf C\langle\mathbf H\rangle)$ with $\mathbf C$ the gauge-projected ensemble covariance and $\langle\mathbf H\rangle$ evaluated at the ensemble mean. Panels D and F used the sweep of Figure~2. The non-Gaussianity of each SGD ensemble was measured from the mode amplitudes of its replicas, $a_\alpha^{(j)}=\mathbf u_\alpha^{\top}\mathbf W^{(j)}\mathbf v_\alpha$. At time $t$, we computed $m_3(t)=\sum_\alpha|\overline{(a_\alpha-\bar a_\alpha)^3}|$, where the bar denotes the average over replicas after discarding the $1\%$ most extreme values of $a_\alpha$ at each end. The value reported for an ensemble is the peak $m_3=\max_t m_3(t)$, evaluated at every $\max(1,\lfloor n/60\rfloor)$-th of the $n$ steps. The ordinate of panel D is $\max_t\mathrm{Tr}\,\mathbf C_{\rm SLF}/\max_t\mathrm{Tr}\,\mathrm{Cov}(\bth)-1$, with the ensemble covariance gauge-projected. Cells with ratio below $1$ are omitted. The line is a least-squares fit over cells with $m_3\geq10^{-2}$. The gray band is an estimate of the finite-$N_s$ floor, spanning $0.5$ to $2$ times the median ordinate over cells with $m_3<10^{-3}$. Open symbols in Figures~2C,F and~3D,F mark cells whose $m_3$ exceeds the value at which the fitted law gives an ordinate of $1$, i.e. where the Gaussian-closure overshoot equals the quantity itself.

\subsection*{Nonlinear network (Figure~\ref{fig:nonlinear})}

\paragraph{Network and data.} Digits $0$ and $1$ from the MNIST training set were used as inputs, with $N=10^4$ balanced examples, downsampled from $28\times28$ to $14\times14$ by $2\times2$ average pooling ($d=196$) and standardised per pixel, and targets were $y=\pm1$. The network output was $\hat y=\vec W_2\mathrm{GELU}(\vec W_1\vec x+\vec b_1)+b_2$ with $h=24$ hidden units, so $P=hd+2h+1=4753$. The $\mathrm{GELU}$ nonlinearity was chosen to ensure that second- and higher-order derivatives for theoretical analysis would be well defined. The per-sample loss was $\ell_i=\frac12(\hat y_i-y_i)^2$, and $L$ was its mean over a minibatch, with no weight decay. All computations were in float64.

\paragraph{Initial conditions and ensembles.} Parameters were initialised at unit scale (seed $0$) and trained by $150$ steps of full-batch gradient descent at learning rate $0.05$, giving a common warm start. For each seed $s$ and each $B$, a parent ensemble of $256$ SGD replicas was run from the warm start for $500$ steps at the production $(\eta,B)$, and one of its replicas, chosen at random, became the initial condition $\bth_J$. Each segment then started all $256$ replicas at $\bth_J$, so $\bC(0)=0$ and the covariance was generated entirely by minibatch noise during the segment. Minibatches were drawn independently for each replica, with replacement. The learning rate was fixed at $\eta=0.13345$; one optimiser step advanced $t$ by $\eta$, so a $200$-step segment spanned $t\approx26.7$.

\paragraph{Integration of the SLF equations.} 
With $D_t=\partial_t+(\eta/2)\partial_t^2$, each equation was integrated as a first-order system, $\dot{\bm}=\mathbf{p}$, $\mathbf{p}=(2/\eta)(\mathbf{R}_m-\mathbf{p})$ with $\mathbf{R}_m$ the right-hand side of \eqref{eq:20260501a}, and likewise for $\bC$, by fixed-step fourth-order Runge--Kutta with two substeps per optimiser step. $\langle\bH\rangle$ was computed exactly at every optimiser step and interpolated linearly within it, as were the source terms of \eqref{eq:20260501b}; the covariance-dependent source terms used $\bC$ from the previous optimiser step, and the correction $\frac12\nabla\mathrm{Tr}(\bC\langle\bH\rangle)$ was held fixed across each step. Initial velocities took their slow-manifold values, $\dot{\bm}(0)=\mathbf{R}_m-(\eta/2)\dot{\mathbf{R}}_m$ and its analogue for $\bC$, so that the spurious fast mode of $D_t$, at rate $2/\eta$, was not excited. In a $50$-step segment at $B=256$, halving the substep changed $\mathrm{Tr}\,\bC_\mathrm{pred}$ by less than one part in $10^4$.

\paragraph{Terms of (24).} 
All terms were evaluated at $\bm(t)$. For minibatches drawn with replacement,
\begin{equation*}
\langle\bH\bC\bH\rangle=\langle\bH\rangle\bC\langle\bH\rangle
+\frac1B\mathbb{E}_i\left[\delta\bH_i\,\bC\,\delta\bH_i\right],
\text{ where } \delta\bH_i=\bH_i-\langle\bH\rangle ,
\end{equation*}
and $\mathrm{Cov}(\mathbf{f})=B^{-1}\mathrm{Cov}_i(\nabla\ell_i)$. The fluctuation term was computed exactly by using the structure of the per-sample Hessian. The output $\hat y$ depended on $\bth$ only through the $2h+1$ scalars $z_j=\mathbf{W}_{1j}\cdot\mathbf{x}+b_{1j}$, $W_{2j}$ and $b_2$, each affine in $\bth$, so $\nabla^2\hat y$ had rank at most $2h$ and $\bH_i=\mathbf{g}_i\mathbf{g}_i^\top+r_i\nabla^2\hat y_i = \mathbf{U}_i\mathbf{D}_i\mathbf{U}_i^\top$, with $\mathbf{g}_i=\nabla\hat y_i$, $r_i$ the residual, $\mathbf{U}_i\in\mathbb{R}^{P\times(2h+1)}$ known in closed form and $\mathbf{D}_i$ block diagonal. The same structure made the two terms involving $\nabla\mathrm{Tr}(\bC\bH)$ exact. Because the $2h+1$ scalars were affine in $\bth$, the directions along which they varied did not depend on $\bth$; writing $\mathrm{Tr}(\bC\bH_i)$ as a function of the scalars, with $\bC$ contracted onto those fixed directions, gave $\mathbf{q}_i=\nabla\mathrm{Tr}(\bC\bH_i)$ by automatic differentiation. Then the correction in \eqref{eq:20260501a} was $\frac12\nabla\mathrm{Tr}(\bC\langle\bH\rangle)=\frac12\mathbb{E}_i[\mathbf{q}_i]$, and the cross covariance in \eqref{eq:20260501b} was $\mathrm{Cov}(\mathbf{f},\nabla\mathrm{Tr}(\bC\bH))=-B^{-1}\mathrm{Cov}_i(\nabla \ell_i,\mathbf{q}_i)$, where $\bH$ in the covariance was the minibatch Hessian. For the measured-covariance prediction the correction was instead averaged over the $256$ replica displacements, which was exact for the sample covariance.

The segments in panels A-C were run with stochastic estimates of these two terms: $256$ Gaussian draws $\vec z\sim N(0,\vec C)$ for the correction, using $\mathbb{E}_{\mathbf{z}}[\mathbf{z}^\top\langle\bH\rangle\mathbf{z}]=\mathrm{Tr}(\bC\langle\bH\rangle)$, and $8$ Gaussian probes for the cross term, with variances those of $\vec C$ along the eigenvectors of $\langle\bH\rangle$ and retained on the $64$ eigenvectors of largest variance. Repeating the $200$-step $B=256$ segment for one seed with the exact evaluation, on identical minibatch streams, changed $e_\mathrm{closed}$ by at most $0.18\%$ at any step, $\mathrm{Tr}\bC_\mathrm{pred}/\mathrm{Tr}\bC_\mathrm{meas}$ by at most $0.07\%$, and the panel-B ratio from $1.1376$ to $1.1373$. All points at $B\le128$ in panel D used the exact
evaluation.

\paragraph{Error measures.} 
Each prediction's error was $e(t)=\|\bm(t)-\mathbb{E}[\bth](t)\|/\|\mathbb{E}[\bth](t)\|$, where $\mathbb{E}[\bth]$ is the empirical ensemble mean, evaluated at every step. Panel B showed, for each segment, $e_\mathrm{no-C}/e_\mathrm{closed}$ at step $200$. This value lay within $2.5\%$ of the segment's median over steps $190$-$200$ in every segment, so it was not a single noisy draw; the median over all $200$ steps gave the same ordering, with medians $1.037$, $1.077$ and $1.134$. The sampling floor was $\|\mathbb{E}_A[\bth]-\mathbb{E}_B[\bth]\|/(2\|\mathbb{E}[\bth]\|)$ for disjoint halves $A$ and $B$ of the ensemble. The covariance traces were $\mathrm{Tr}\,\bC_\mathrm{meas}=n^{-1}\sum_j\|\bth_j-\mathbb{E}[\bth]\|^2$ and the trace of the integrated $\bC$; the unbiased normalisation $1/(n-1)$ would lower every trace ratio in panels C and D by $0.39\%$. The sampling error of $\mathrm{Tr} \,\bC_\mathrm{meas}$ was the RMS over $200$ random $128/128$ partitions of $\frac12|\mathrm{Tr}\,\bC_A-\mathrm{Tr}\,\bC_B|$, relative to $\mathrm{Tr}\,\bC_\mathrm{meas}$, computed from the replica ensembles stored for panel D (below); the $1.3\%$ quoted for panel C was its median over the states at $B\ge256$. A replica whose parameters became non-finite was removed, and ensemble statistics were taken over the remainder.

\paragraph{Gauge.} 
The loss was invariant under permutations of hidden units, so $\mathbb{E}[\bth]$ was a meaningful network only if the replicas shared one assignment of units. Aligning each replica to a consensus reference required a non-identity permutation for none of the $2560$ replicas examined at each $B\ge128$. GELU is not odd, so there was no sign symmetry to check.

\paragraph{Panel D and the breakdown at small $B$.} 
The segments at $B\in\{32,64,128\}$ used the same protocol with seed $11$. Replica ensembles were stored at $10$ times during the first $50$ steps at each of the six batch sizes. At each stored state, the tail index was the Hill estimator applied to the upper $10\%$ of the distribution of $L(\bth_j)$ across replicas, checked against a Gaussian surrogate with the same mean and covariance, and the trace error was read at the same step. At $B\le128$, $\mathrm{Tr}\,\bC_\mathrm{pred}$ came from repeating the $50$-step segment with the exact evaluation described above; the replicas, and hence $\mathrm{Tr}\,\bC_\mathrm{meas}$ and the tail index, were identical between the two runs. Seven states, all at $B=32$, were excluded because $\mathrm{Tr}\,\bC_\mathrm{meas}$ or $\mathrm{Tr}\,\bC_\mathrm{pred}$ was non-finite: the ensemble had diverged, and the eigendecomposition used in evaluating the fluctuation term failed. Indefiniteness of the integrated covariance was measured by its smallest eigenvalue divided by its largest. Where $\mathrm{Tr}\,\bC_\mathrm{meas}$ had diverged, $\mathrm{Tr}\,\bC_\mathrm{pred} / \mathrm{Tr}\,\bC_\mathrm{meas}$ was negligible and the plotted error saturated at $1$. The horizontal lines were the median, over each batch size's states, of the sampling error of $\mathrm{Tr}\,\bC_\mathrm{meas}$ at that state; a state was counted as exceeding its sampling error by comparison with its own value. That sampling error required the loss to have finite variance across replicas, so it was not defined below a tail index of $2$, and the lines at $B=64$ and $B=32$ indicated only its scale.

\subsection*{Edge-of-stability threshold test (Figure \ref{fig:eos})}

\paragraph{Network and task.} The network was a two-layer perceptron, $\hat y=\mathbf W_2\tanh(\mathbf W_1\mathbf x+\mathbf b_1)+b_2$, with $d_x=15$ inputs, $h=32$ hidden units and a scalar output ($P=545$), trained on the loss $L=\frac{1}{N}\sum_{i=1}^N\frac{1}{2}(\hat y_i-y_i)^2$ over $N=200$ fixed training examples. The inputs were whitened exactly to a power-law covariance: with $\mu_k\propto k^{-1}$ normalized to unit mean, the input matrix was $\mathbf X=\sqrt N\mathbf Q\,\mathrm{diag}(\mu_k^{1/2})$, where $\mathbf Q$ is the orthonormal factor of a $200\times15$ standard Gaussian matrix. Targets were produced by a random teacher of the same form with $12$ hidden units (weights drawn from $\mathcal N(0,1/\text{fan-in})$, hidden biases from $\mathcal N(0,0.3^2)$, output bias zero), standardized to zero mean and unit variance and multiplied by $4$. The student was initialized with weights drawn from $\mathcal N(0,0.25/\text{fan-in})$ and zero biases. The learning rate was $\eta=0.1$, so $2/\eta=20$. 

\paragraph{Base points.} Full-batch gradient descent was run for $1500$ steps from this initialization. At each step $k\geq1$ we formed the bounce average $\mathbf m_k=\frac{1}{2}(\bth_{k-1}+\bth_k)$ and computed $\lambda_1$ and $\mathbf v_1$ of the full-batch Hessian by Lanczos iteration. The interval $[0.63,1.10]\times2/\eta$, truncated at the largest $\lambda_1$ reached, was divided into $44$ equal bins in $\lambda_1$. From each non-empty bin, the middle (in time) of the steps falling in that bin was taken as a base point, giving $34$ base points between steps $12$ and $1377$ with $\lambda_1$ from $13.31$ to $21.47$. At each base point we computed the per-sample curvatures $h_i=\mathbf v_1^{\top}\mathbf H_i\mathbf v_1$, $\kappa=\mathrm{Var}_i(h_i)/\lambda_1^2$ and $d$ of \eqref{eq:20260807g}, by automatic differentiation.

\paragraph{Probe.} At each base point $\mathbf m$ and each $B$, we displaced $n=6\times10^4$ antithetic pairs of replicas to $\mathbf m\pm\epsilon\mathbf v_1$, with $\epsilon=10^{-5}|\mathbf m|$. Both members of a pair took one SGD step with the same batch of $B$ examples, drawn uniformly with replacement, and different pairs used independent batches. The half difference of their displacements along $\mathbf v_1$, $p=\frac{1}{2}\mathbf v_1^{\top}(\bth'_+-\bth'_-)$, cancels the gradient-noise source exactly and gives the mean-square growth rate $r=\eta^{-1}\ln(\overline{p^2}/\epsilon^2)$, whose zero is $\lambda_0$ with $d=0$. The standard error of $r$ was propagated from the sample variance of $p^2$. The measured threshold at each $B$ was the largest $\lambda_1$ at which $r$, ordered by $\lambda_1$, changed sign, located by linear interpolation between the two bracketing base points. Its error combined the standard errors of $r$ at those points, divided by the slope of the interpolant. This error was then multiplied by $1.710$, the root mean square of the normalized differences between these thresholds and those from a repeat of the sweep at $B=5$--$128$ with different random draws.

\paragraph{Prediction and comparison.} 
The predicted threshold was $\lambda_0(d{=}0)=2/[\eta(1+\kappa/B)]$, with $\kappa$ linearly interpolated in $\lambda_1$ between base points and evaluated at the measured threshold. The curve including the second term of \eqref{eq:20260807j} subtracts $\eta d/(2B)$, with $d$ interpolated in the same way. Residuals are $100(\lambda_0^{\rm pred}-\lambda_0^{\rm probe})/\lambda_0^{\rm probe}$. The reported $\chi^2/\mathrm{dof}$ is the mean of the squared normalized residuals over the seven batch sizes, with no fitted parameters.

\subsection*{Use of AI tools}
The theoretical framework, the derivation of the SLF equations, and the design of the study are the authors’ own. In developing the numerical experiments and the manuscript, we made use of AI assistants based on Anthropic’s Claude model Opus 5.5. AI coding agents wrote and ran simulation and analysis code for the DLN and nonlinear-network experiments under specifications made by the authors. An AI assistant also helped interpret results, check derivations, survey related literature, and edit parts of the text. The authors reviewed, verified, and revised all AI-generated code, analyses, and text, and take full responsibility for the content of the paper.

\pagebreak

\section*{Supplemental Information}

\setcounter{figure}{0} 
\renewcommand{\thefigure}{S\arabic{figure}} 

\begin{figure}[h]
    \centering
    \includegraphics[width=0.5\linewidth]{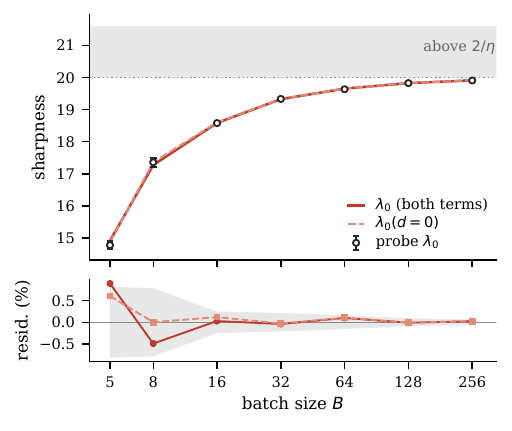}
    \caption{The edge of stability is shifted by batch noise. Mean-square EoS threshold $\lambda_0$ calculated in a two-layer network from \eqref{eq:20260807j}, either with (solid line) or without (dashed line) the second term from the equation. Points show critical curvature based on a one-step probe estimator in the same network. Lower panel shows the relative error between the measured and calculated values. Error bars in the top panel and shading in the lower panel show $\pm 2$ std of the probe measurement.
    }
    \label{fig:eos}
\end{figure}

\end{document}